\def\year{2022}\relax
\documentclass[letterpaper]{article} % DO NOT CHANGE THIS
\usepackage{aaai22}  % DO NOT CHANGE THIS
\usepackage{times}  % DO NOT CHANGE THIS
\usepackage{helvet}  % DO NOT CHANGE THIS
\usepackage{courier}  % DO NOT CHANGE THIS
\usepackage[hyphens]{url}  % DO NOT CHANGE THIS
\usepackage{graphicx} % DO NOT CHANGE THIS
\usepackage{natbib}  % DO NOT CHANGE THIS AND DO NOT ADD ANY OPTIONS TO IT
\usepackage{caption} % DO NOT CHANGE THIS AND DO NOT ADD ANY OPTIONS TO IT
\DeclareCaptionStyle{ruled}{labelfont=normalfont,labelsep=colon,strut=off} % DO NOT CHANGE THIS
\usepackage{amsmath}
\usepackage{amssymb}

\usepackage{algorithm}
\usepackage{algorithmic}
\usepackage{amsthm}
\usepackage{amsfonts}       % blackboard math symbols
\usepackage{nicefrac}       % compact symbols for 1/2, etc.
\usepackage{booktabs}
\usepackage{multirow}
\usepackage{multicol}
\usepackage{microtype} 
\usepackage{tabularx}
\usepackage{array}
\usepackage{xcolor}
\usepackage{dsfont}
\usepackage{pifont}% http://ctan.org/pkg/pifont
\newcommand{\cmark}{\ding{51}}%
\usepackage{etoolbox}
\preto\tabular{\setcounter{magicrownumbers}{0}}
\newcounter{magicrownumbers}
\newcommand\rownumber{\stepcounter{magicrownumbers}\arabic{magicrownumbers}}

\newtheorem{theorem}{Theorem}
\theoremstyle{definition}
\newtheorem{observation}[theorem]{Observation}
\theoremstyle{definition}
\newtheorem*{remark}{Remark}

\usepackage{newfloat}
\usepackage{listings}
\floatstyle{ruled}
\newfloat{listing}{tb}{lst}{}
\floatname{listing}{Listing}
\title{Static-Dynamic Co-Teaching for Class-Incremental 3D Object Detection}
\author{
Na Zhao~~~Gim Hee Lee
}
\affiliations{
Department of Computer Science, National University of Singapore\\
\{nazhao, gimhee.lee\}@comp.nus.edu.sg
}

\usepackage{bibentry}
\begin{document}

\maketitle

\begin{abstract}
Deep learning-based approaches have shown remarkable performance in the 3D object detection task. However, they suffer from a catastrophic performance drop on the originally trained classes when incrementally learning new classes without revisiting the old data. This ``catastrophic forgetting'' phenomenon impedes the deployment of 3D object detection approaches in real-world scenarios, where continuous learning systems are needed. In this paper, we study the unexplored yet important class-incremental 3D object detection problem and present the first solution - SDCoT, a novel static-dynamic co-teaching method. Our SDCoT alleviates the catastrophic forgetting of old classes via a static teacher, which provides pseudo annotations for old classes in the new samples and regularizes the current model by extracting previous knowledge with a distillation loss. At the same time, SDCoT consistently learns the underlying knowledge from new data via a dynamic teacher. 
%The static teacher is the previous model trained with old data, while the dynamic teacher is an ensemble of current model across up-to-date training steps. 
We conduct extensive experiments on two benchmark datasets and demonstrate the superior performance of our SDCoT over baseline approaches in several incremental learning scenarios.
\end{abstract}

\section{Introduction}
\begin{figure}[t]
\centering
\includegraphics[scale=0.4]{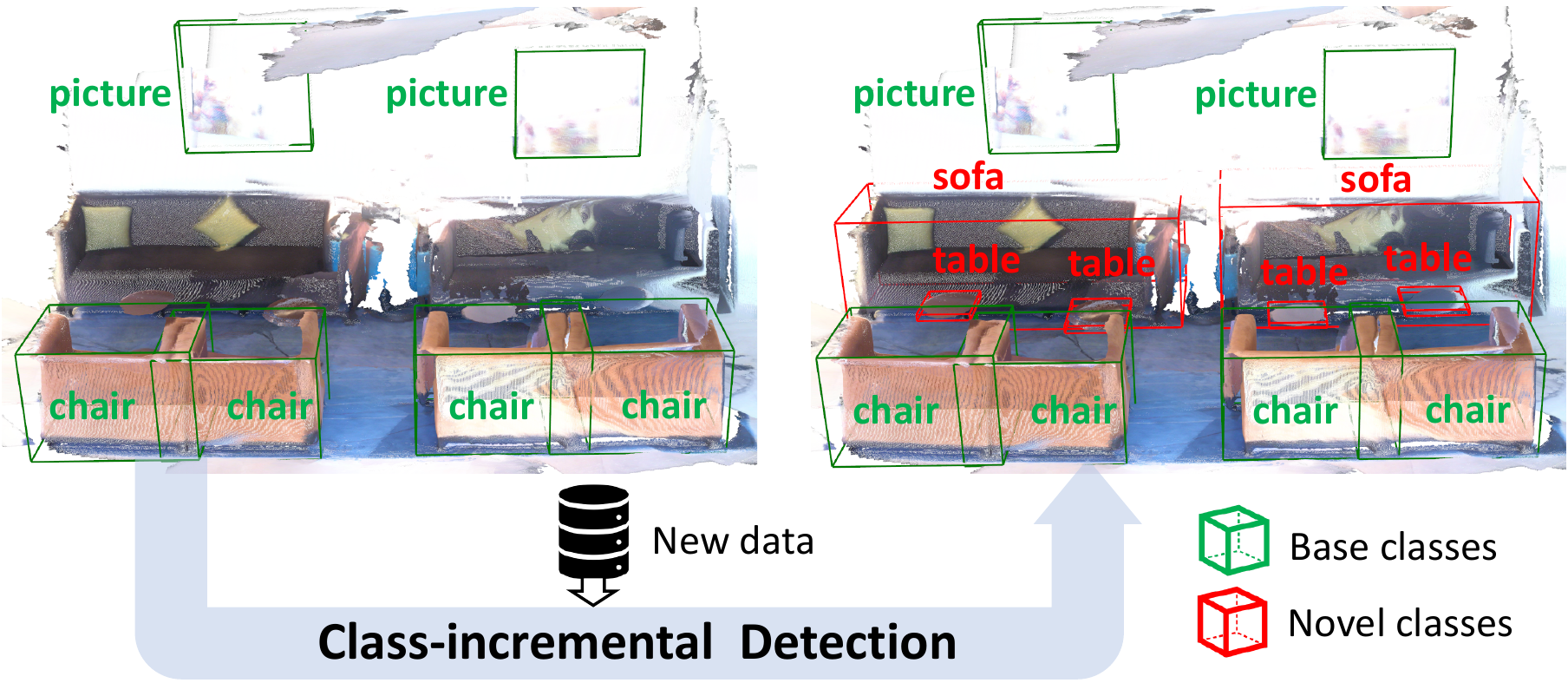}
\vspace{-0.1in}
\caption{\small{An example of class-incremental 3D object detection.}}
\vspace{-0.2in}
\label{fig:teaser}
\end{figure}

The success of deep learning %has been witnessed 
are seen in many computer vision tasks that include point cloud-based 3D object detection. Many deep learning-based approaches \cite{li2016vehicle, chen2017multi, beltran2018birdnet, yan2018second, yang2018pixor, zeng2018rt3d, zhou2018voxelnet, chen2019fast, lang2019pointpillars, qi2019deep, shi2019pointrcnn, yang2019std, zhou2019fvnet, yang20203dssd, zheng2021cia} are proposed and have shown impressive performance in localizing and categorizing objects of interest in the point cloud of a scene. 
However, these approaches suffer from ``\textit{catastrophic forgetting}'', \textit{i.e.} a significant performance degradation on the old classes (\textit{c.f.} Row 3 of Table~\ref{tab:sunrgbd_batch} and \ref{tab:scannet_batch}) when applied in a class-incremental scenario where new classes are added incrementally while old data might be unavailable due to storage limitation or privacy issue. 
The ``catastrophic forgetting'' phenomenon largely limits the use of these models in real-world applications, where intelligent machines are required to continually learn new knowledge without forgetting the old one.
%However, these approaches suffer from the ``\textit{catastrophic forgetting}'' phenomenon, \textit{i.e.} a significant performance degradation on the old classes when trained on data of new classes that are added incrementally without revisiting the old data. These old data might be unavailable due to storage limitation or privacy issue. 
%The ``catastrophic forgetting'' phenomenon largely limits the use of these models in real-world applications, where intelligent machines are required to incrementally learn new tasks with new data.
For example, the detection system on a domestic robot is initially trained to detect several base classes such as `chair' and `picture' (see the \textit{left} example in Figure \ref{fig:teaser}).
Subsequently, when the examples of novel classes such as `sofa' and `table' become available, the system needs to incrementally learn to detect these novel classes without losing the ability to detect the base classes (see the \textit{right} example in Figure \ref{fig:teaser}).
Furthermore, the ability to do \textit{class-incremental learning} of 3D object detection gives machines a learning capability closer to humans since we do not forget old concepts after learning new ones.

Although class-incremental learning has been studied in several computer vision tasks \cite{li2017learning, shmelkov2017incremental, michieli2019incremental, dong2021i3dol}, especially image classification, class-incremental learning of 3D object detection remains unexplored. 
To our best knowledge, we are \textit{the first} to study this unexplored yet important problem, and to present an effective Static-Dynamic Co-Teaching solution named SDCoT.
Our SDCoT is able to incrementally detect new classes \textit{without} revisiting any old data or annotations of the old classes. 
A challenge in class-increment learning of object detection is the high chance of old (in the background without labels) and new (with labels) classes co-occurring in the new training samples.  
This causes the model to wrongly suppress the old classes and thus expedites the catastrophic forgetting process. 
To overcome this challenge, SDCoT leverages the previous model trained on old data to generate pseudo annotations of old classes in the new training samples. Consequently, a mixture of pseudo labels of the old classes and the ground-truth labels of new classes, \textit{i.e.} ``mixed labels'' is used to train the current model.

A naive way of pseudo label generation leads to inaccurate and incomplete pseudo labels that deteriorate the detection performance. 
Our SDCoT alleviates this problem by introducing co-teaching from two teachers: a static teacher and a dynamic teacher. Specifically, the static teacher is a frozen copy of the previous model, which teaches to distill previously learned knowledge from \textit{old data} with a distillation loss. On the other hand, the dynamic teacher is an ensemble of the current model across its up-to-date training steps, which teaches to consistently learn the underlying knowledge from the \textit{new data} with a consistency loss. As a result, our SDCoT trains the current model with supervision from the ``mixed labels'' and regularizations from the two adversarial teachers.
We conduct extensive experiments on SUN RGB-D and ScanNet datasets. The performance improvements over the baselines under different incremental learning scenarios demonstrate the effectiveness of our SDCoT in class-incremental 3D object detection. Additionally, we validate the contribution of static and dynamic teachers in knowledge exploitation by evaluating different variants of our SDCoT. Finally, we also show our SDCoT is compatible with examples from old data once they are available.

% To summarize, we make the following contributions: 
% \begin{itemize}
%     \vspace{-0.1in}
%     \item We are the first to study the unexplored yet important class-incremental 3D object detection problem.
%     \vspace{-0.1in}
%     \item We propose a novel and effective static-dynamic co-teaching method to solve this challenging problem.
%     \vspace{-0.1in}
%     \item We carry out extensive experiments on two benchmark datasets and demonstrate the effectiveness of our SDCoT in a variety of incremental learning scenarios.
% \end{itemize}

\section{Related Work}\label{sec:related_work}
%\paragraph{3D Object Detection.}
%The task of point cloud based 3D object detection is to identify all the interested objects and localize them with oriented 3D bounding boxes in the point cloud of a scene. Due to the irregular and sparsity characteristics of 3D point clouds, it is challenging to localize 3D objects in an efficient way like its image-based counterparts. Several prior works \cite{qi2019deep, shi2019pointrcnn, yang20203dssd, yang2019std} overcome this challenge by exploring the sparsity of 3D data and generating 3D proposals around a set of seed points that are determined by using various techniques, \textit{e.g.} segmenting~\cite{shi2019pointrcnn, yang2019std} or voting~\cite{qi2019deep, yang20203dssd}). Despite the promising results achieved by these works, they suffer from severe performance degradation, \textit{i.e.} catastrophic forgetting when required to incrementally learn new classes without the availability of the old data. This is because these approaches that are trained offline with predefined classes do not have the capacity to learn incrementally. In this paper, we address this largely overlooked, yet important class-incremental 3D object detection problem.

Class-incremental learning is a classical machine learning problem, which refers to the continuous addition of new classes into a model. 
Most existing class-incremental leaning methods focus on image classification task, which can be classified into two main categories:
1) \textit{regularization based methods} %\cite{kirkpatrick2017overcoming, li2017learning, aljundi2018memory, hou2019learning} 
minimize the discrepancy between either the data \cite{li2017learning, hou2019learning} or parameters \cite{kirkpatrick2017overcoming, aljundi2018memory} of the previous model and the current model; 
2) \textit{rehearsal/replay-based methods} 
%\cite{castro2018end, ostapenko2019learning, rebuffi2017icarl, shin2017continual, wu2019large} 
store a subset of exemplars from previous classes \cite{rebuffi2017icarl, castro2018end, wu2019large} or produce synthesized exemplars for previous classes using a generative model \cite{shin2017continual, ostapenko2019learning}. 
 
Recently, several works apply class-incremental learning on image-based object detection task. 
Most of them \cite{shmelkov2017incremental, chen2019new, hao2019end, peng2020faster} address this problem by exploring knowledge distillation on network responses (\textit{i.e.} data-based regularization). 
For example, the first study on class-incremental image object detection \cite{shmelkov2017incremental} leverages Fast R-CNN as object detector and applies distillation losses on the predictions of classification layer and bounding box regression layer. Built upon this first work, CIFRCN \cite{hao2019end} additionally distills the intermediate features of RPN by adopting Faster R-CNN.
However, these knowledge distillation methods are specifically designed for 2D object detection backbones; how to apply knowledge distillation (\textit{e.g.} what to distill) on the point cloud-based 3D object detection backbones is unknown. 
% We contribute to the adaptation of our 3D object detector backbone and show the effects of different choices in employing knowledge distillation.
We adapt a standard 3D object detector to class-incremental 3D object detection task, and further show the effects of different choices in employing knowledge distillation on adapted 3D object detector. More recently, IncDet \cite{liu2020incdet} adapts Elastic weight consolidation (EWC) \cite{kirkpatrick2017overcoming}, a parameter-based regularization method, to class-incremental image object detection task. 
IncDet circumvents the co-occurrence challenge in class-incremental object detection by using pseudo bounding box annotations of old classes in new training samples. 
Similar to IncDet, we also utilize pseudo annotations of old classes to prevent the current model from mistakenly classifying old class objects as background in the new samples. Nonetheless, unlike its image-based counterpart, the generated pseudo annotations in 3D scenario are not very accurate and may cause performance degradation. We solve this issue by proposing a static-dynamic co-teaching technique. 

%In this paper, we aim to do class-incremental 3D object detection in a challenging scenario, where \textit{no} old data is available during class-incremental learning. In view of this goal, the category of regularization-based methods is a better choice than replayed-based methods. 
%Parameter-based regularization aims to restrain weight drift of important parameters in the previous model when fine-tuning on new data sets. However, it needs to additionally estimate the importance of each parameter in the previous model, which is not trivial to determine. In contrast, data-based regularization only restrains activation drift of either intermediate or final outputs from the previous model via knowledge distillation \cite{hinton2015distilling}, which is simple but effective.
%Inspired by the data-regularization, we leverage knowledge distillation to transfer knowledge from the previous 3D object detector to the current detector.

\section{Our Methodology}
\subsection{Problem Definition}
 In the class-incremental 3D object detection task, there are two non-overlapped sets of classes: \textit{base classes} set $C_{base}$ and \textit{novel classes} set $C_{novel}$. A set of data $D_{base}$ is available for $C_{base}$, and another set of data $D_{novel}$ is available for $C_{novel}$. We define the \textbf{class-incremental 3D object detection} task as follows: given a well-trained 3D object detector $\Phi_B$ (\textit{i.e.} base model) on $D_{base}$, our goal is to learn an incremental 3D object detector $\Phi_{B \cup N}$ (\textit{i.e.} incremental model) using only $D_{novel}$, such that $\Phi_{B \cup N}$ is able to detect objects from all the classes seen so far, \textit{i.e.} $C_{base} \cup C_{novel}$. 
To this end, we propose SDCoT: a novel Static-Dynamic Co-Teaching framework to achieve class-incremental learning on 3D object detection.

\begin{figure}[t]
	\centering
	\includegraphics[scale=0.32]{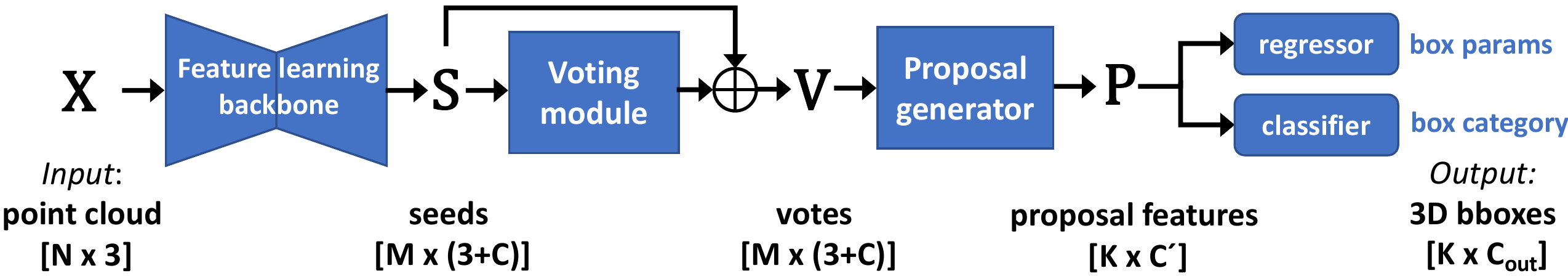}
	\vspace{-0.1in}
	\caption{\small{Overview of our backbone - modified VoteNet.}}
    \vspace{-0.2in}
	\label{fig:votenet_framework}
\end{figure}

\subsection{Anatomy of VoteNet} \label{sec:AnatomyVoteNet}
We use VoteNet~\cite{qi2019deep} as the prototype of our 3D object detector because of its efficiency and simplicity in point cloud-based 3D object detection. In this section, we dissect the anatomy of VoteNet to reveal two observations that we leverage to adapt VoteNet for the design of our SDCoT.

\vspace{-0.05in}
\begin{observation} \label{obs:votenetSamples}
VoteNet inherently includes two sub-sampling steps: 1) sub-sample $M$ seeds (denoted as $\textbf{S}$ in Figure \ref{fig:votenet_framework}) from $N$ input points via a feature learning backbone; and 2) sub-sample $K$ votes from $\textbf{V}$ as cluster centers to generate $K$ proposals by aggregating neighboring votes. Due to the stochasticity of these sub-sampling steps in VoteNet, different sets of proposals are produced from the same input point cloud at different times.
\end{observation}

\vspace{-0.1in}
\begin{remark}
The stochasticity of VoteNet implies that the sets of proposals generated from the base and the incremental models, respectively, are not aligned even for the same input point cloud. This impedes a direct comparison of the proposals, which is essential for training an incremental model via knowledge distillation. To circumvent this problem, we store all the indices of the sampled points and the indices of the sampled votes from the incremental model, and re-use these indices in the base model. Consequently, the two sets of proposals produced from the two models are aligned and can be compared to measure the output discrepancy.
\end{remark} 

\vspace{-0.1in}
\begin{observation} \label{obs:votenetFixPredSize}
After obtaining the proposal features (denoted as $\textbf{P}$ in Figure \ref{fig:votenet_framework}), VoteNet adopts one multi-layer perceptron (MLP) layer to yield prediction scores for each proposal. The prediction scores consist of 2 objectness scores, 3 center offsets, $2NH$ heading scores ($NH$ heading bins), $4NC$ box size scores ($NC$ size templates), and $NC$ category scores. Note that the box size scores include 1 classification score and 3 size offsets for each size template, and the size templates correspond to the class categories.  
The size of prediction scores is fixed after VoteNet is trained.
\end{observation}

\vspace{-0.1in}
\begin{remark}
The fixed prediction scores size of VoteNet after training is problematic for class-incremental learning. To enroll new classes in class-incremental learning, the weights for class-aware predictions need to be dynamically updated according to the addition of novel classes. 
We solve this problem by first decoupling the last MLP layer into two parts (\textit{i.e.} regressor and classifier in Figure \ref{fig:votenet_framework}) to separate the category prediction from the predictions of other scores, and then adding new weights to the classifier according to the novel classes.
We concurrently replace the class-aware size prediction with class-agnostic one to achieve a simpler implementation for class-incremental 3D object detection.
\end{remark}

\subsection{Our SDCoT}
\subsubsection{Pseudo Label Generation.}\label{sec:pseudolabel}
A challenge in class-incremental learning of object detection is the high possibility of co-occurrence of different classes in some scenes. Concretely, there is a high probability that instances belonging to the base classes appear as background in the samples of $D_{novel}$. 
As a result, these regions that contain the old class objects are wrongly suppressed during incremental class training and thus expedite catastrophic forgetting. Moreover, the presence of base classes without annotations confuses the incremental learning model.

\begin{figure}[t]
	\centering
	\includegraphics[scale=0.4]{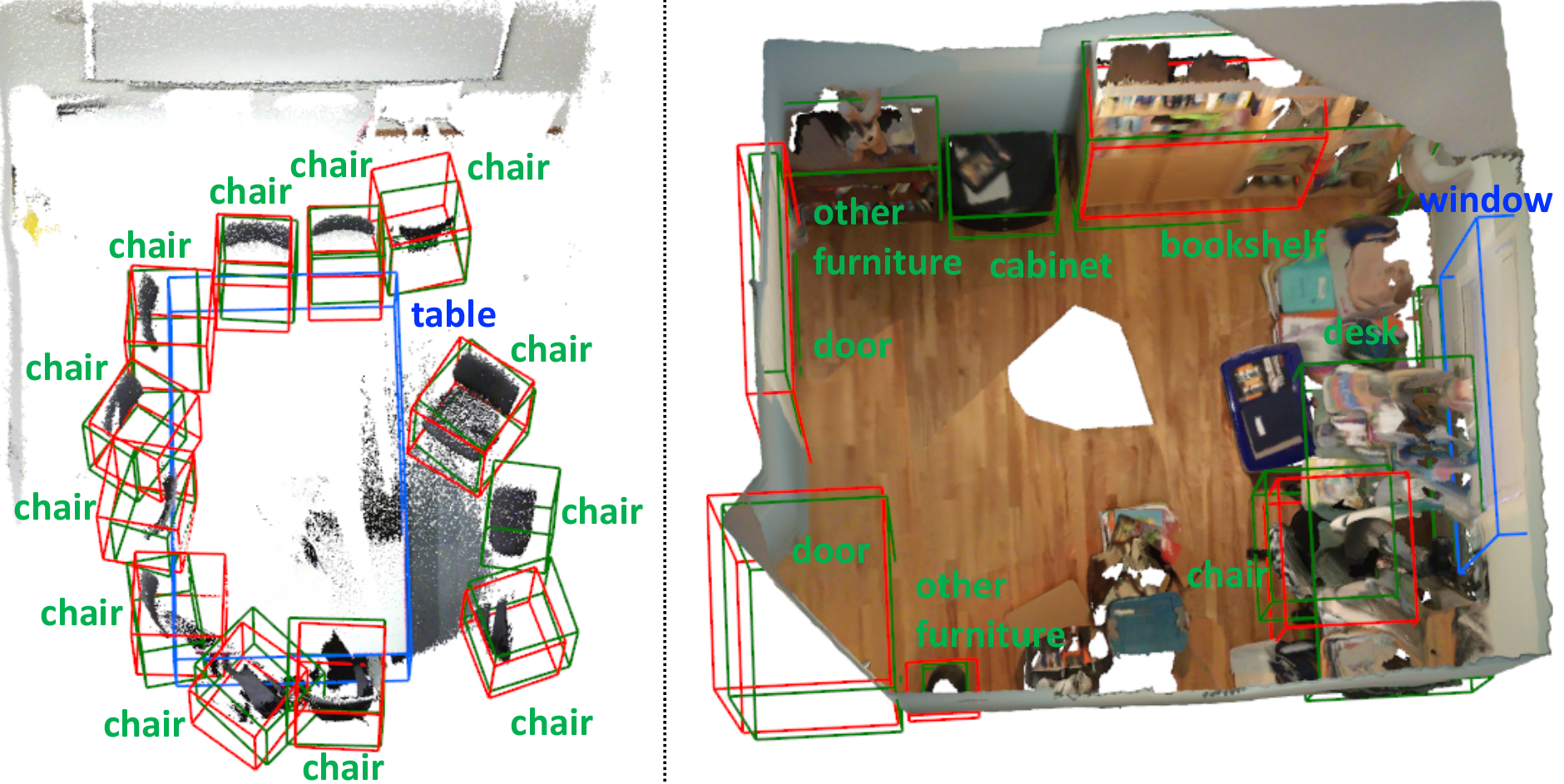}
	    \vspace{-0.1in}
	\caption{\small{Example of generated pseudo labels from SUN RGB-D (left) and ScanNet (right). \textcolor{red}{Red} bboxes are generated pseudo annotations w.r.t $C_{base}$}. \textcolor{green}{Green} and \textcolor{blue}{Blue} bboxes are GT annotations w.r.t $C_{base}$ and $C_{novel}$, respectively.}
	\label{fig:pseudo_labels}
    \vspace{-0.1in}
\end{figure} 

\begin{figure*}[h]
	\centering
	\includegraphics[scale=0.4]{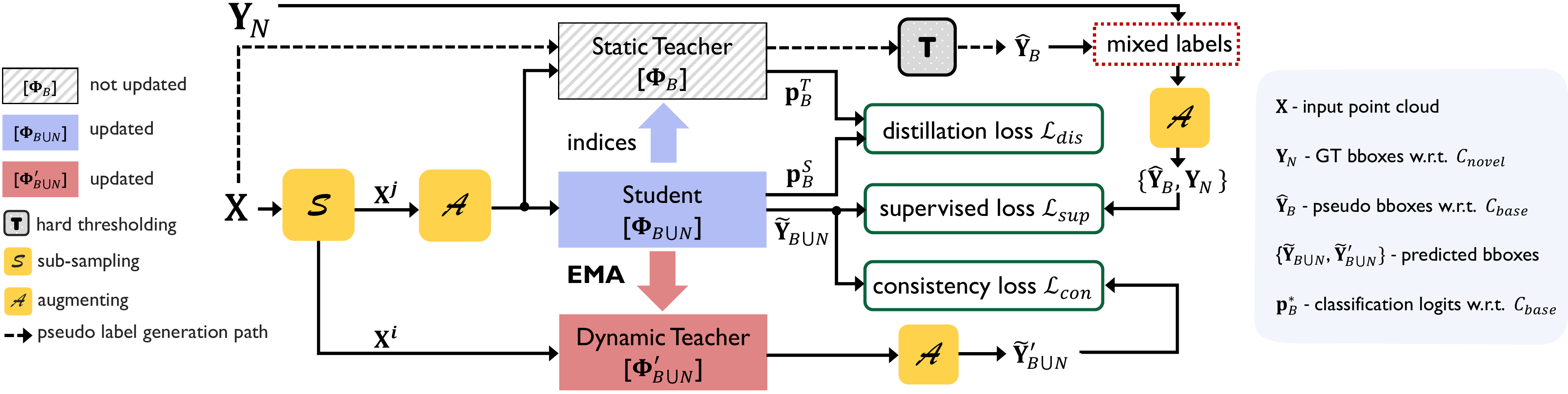}
    \vspace{-0.1in}
	\caption{\small{\textbf{The architecture of our SDCoT}. The student and two teacher networks are 3D object detectors based on modified VoteNet.}} 
	\label{fig:SDCoT_framework}
	\vspace{-0.15in}
\end{figure*}

To overcome the co-occurrence challenge, we take a frozen copy of the base model $\Phi_B$ to generate pseudo labels with respect to $C_{base}$ for each training sample in $D_{novel}$. The generation of pseudo labels from $\Phi_B$ can also be considered as a way to exploit previous knowledge. More specifically, after obtaining the predicted 3D bounding boxes (bboxes) from $\Phi_B$, we filter out low-confidence bboxes by setting two thresholds with respect to the objectness score and classification probability, denoted as $\tau_o$ and $\tau_c$. Unfortunately, the resulting pseudo labels with the hard thresholding strategy are often inaccurate and incomplete, \textit{i.e.} there are missing annotations for some objects of base classes (see examples in Figure \ref{fig:pseudo_labels}).
Consequently, these inaccurate and incomplete labels can affect the learning of the incremental model. 
We alleviate the detrimental effects of these labels by a static-dynamic co-teaching strategy.

\subsubsection{Static-Dynamic Co-Teaching.}\label{sec:static-dynamic_co-teaching}
We design our static-dynamic co-teaching strategy based on the conjecture that the incremental model is less susceptible to noisy and incomplete labels when it is able to largely exploit the underlying knowledge from the base model and new data. 
Generally, the well-trained base model encodes valuable knowledge of base classes. In view of this, we adopt a frozen copy of the base model as our \textbf{static teacher}.
Through the use of pseudo labels, we impede the catastrophic forgetting of base classes caused by the absence of base class annotations in novel training samples. To further exploit more knowledge from the base model, we introduce a distillation scheme with the aim of keeping responses from the base and incremental models to be as close as possible. Specifically, our distillation scheme targets the predicting layer and computes a distillation loss that measures the difference between the classification logits with respect to $C_{base}$ from the base and incremental models.
This knowledge distillation scheme can compensate for the missing labels with respect to $C_{base}$ when the base class objects co-occur in a scene of $D_{novel}$. Furthermore, the responses, \textit{i.e.} classification logits with respect to $C_{base}$ can provide some useful information of the background, \textit{i.e.} dark knowledge \cite{furlanello2018born, hinton2015distilling}, even when there is no base class object.

To exploit more information from the new data, we also design a \textbf{dynamic teacher} that is able to consistently learn the underlying knowledge in terms of both base and novel classes.
The design of our dynamic teacher is inspired by Mean Teacher \cite{tarvainen2017mean}: a self-ensembling technique that is originally proposed to effectively exploit unlabeled data for reducing over-fitting in semi-supervised learning. SESS \cite{zhao2020sess} adapts this self-ensembling technique to semi-supervised 3D object detection task by proposing a perturbation scheme and a consistency loss that enforces the consensus of locations, sizes, categories of the output proposals between a student and a teacher network. More importantly, they show that their superior performance under 100\% labeled data 
is due to the consistency regularization of the mean-teacher paradigm, which gives their framework the capability to exploit additional underlying knowledge from the data. 
Thus, we incorporate the dynamic teacher, and adopt the perturbation scheme and the consistency loss of SESS in our SDCoT for a deeper knowledge exploitation of the new data. Consequently, the dynamic teacher guides the incremental model to be more robust against imperfect pseudo labels in new data and also concurrently to be more expressive on new classes.

\subsubsection{SDCoT Details.} \label{sec:sdcot_overview}
The architecture of our SDCoT is illustrated in Figure \ref{fig:SDCoT_framework}. It consists of three networks: one student, one static teacher, and one dynamic teacher. Both the student and two teacher networks are 3D object detectors that use the modified VoteNet as backbone. Particularly, the student is the incremental detector $\Phi_{B \cup N}$ that incrementally learns from $C_{novel}$. It is co-taught by the static and dynamic teachers. 
The static teacher is a frozen copy of the base model $\Phi_B$, which is used to generate pseudo labels for objects of $C_{base}$ in $D_{novel}$ and prevent the incremental model from drifting too much away from the base model.
The dynamic teacher network $\Phi'_{B \cup N}$ is an exponential moving average of the student network, which dynamically generates targets of all classes for the student network.  
Note that the parameters of the student and dynamic teacher networks are initialized from $\Phi_B$, with the exception that the added weights in the classifier for novel classes are randomly initialized. 

Given an input point cloud denoted as $\mathbf{X}$ in Figure \ref{fig:SDCoT_framework}, our SDCoT first forwards it to the static teacher to generate $K$ 3D bounding boxes (\textit{i.e.} proposals) for the base classes. A subset of these 3D bboxes are selected as pseudo labels $\hat{\mathbf{Y}}_B$ by thresholding with $\tau_o$ and $\tau_c$. The pseudo labels $\hat{\mathbf{Y}}_B$ are combined with the ground-truth labels of novel classes $\mathbf{Y}_N$ to form ``mix labels''.
 
Concurrently, SDCoT sub-samples the input point cloud twice to get two point clouds, \textit{i.e.} $\mathbf{X}^i$ and $\mathbf{X}^j$ in Figure \ref{fig:SDCoT_framework}.  
$\mathbf{X}^i$ is directly passed to the dynamic teacher network, while $\mathbf{X}^j$ is further augmented before inputting into the student and static teacher networks. 
The sub-sampling and augmentation (\textit{i.e.} stochastic flipping, rotation, and scaling) are components of the perturbation scheme, which allows the model to learn useful knowledge rather than memorizing the training data. 

As discussed in Observation~\ref{obs:votenetSamples}, the two stochastic sampling steps (\textit{i.e.} the sampling of seeds and votes) cause the base and incremental models to give unaligned proposals despite the same input. 
We overcome this problem by re-using the selected indices yielded by the student in the static teacher network.
%To overcome this problem, the selected indices yielded by the student network are re-used by the static teacher network. Consequently, the output object proposals of these two networks are aligned. 
A \textbf{distillation loss} $\mathcal{L}_{dis}$ is computed to measure the discrepancy between the classification logits of the proposals from the static teacher (\textit{i.e.} $\mathbf{p}_B^T$ in Figure \ref{fig:SDCoT_framework}) and the logits corresponding to $C_{base}$ from the student (\textit{i.e.} $\mathbf{p}_B^S$ in Figure \ref{fig:SDCoT_framework}). We normalize the classification logits by subtracting its mean over class dimension, which yield $\mathbf{\bar{p}}_{B}^T$ and $\mathbf{\bar{p}}_{B}^S$, respectively. More formally, the distillation loss is computed as: 
   \vspace{-0.05in}
\begin{equation}\label{eq:distillation_loss}
    \mathcal{L}_{dis} = \frac{1}{K} \sum_{i=1}^K || \mathbf{\bar{p}}_{B,i}^S - \mathbf{\bar{p}}_{B,i}^T ||_2.
    \vspace{-0.05in}
\end{equation}
$\mathbf{\bar{p}}_{B,i}^*$ is a $|C_{base}|$-dimensional vector, which represents normalized classification logit of $i$-th 3D object proposal.
On the other hand, the output proposals of the student network (\textit{i.e.} $\tilde{\mathbf{Y}}_{B \cup N}$ in Figure \ref{fig:SDCoT_framework}) are compared with: 1) the mixed labels $\{\hat{\mathbf{Y}}_B, \mathbf{Y}_N\}$ transformed by the same augmentation step that is applied on $\mathbf{X}^j$ to compute a \textbf{supervised loss} $\mathcal{L}_{sup}$ \footnote{The details of $\mathcal{L}_{sup}$ are provided in the supplementary material.}, similar as the multi-task loss in VoteNet; and 2) the output proposals of the dynamic teacher network $\tilde{\mathbf{Y}}'_{B \cup N}$ transformed by the same augmentation step as above to compute a \textbf{consistency loss} $\mathcal{L}_{con}$ as in SESS, respectively.

At each training iteration $t$, the student network is updated by the stochastic gradient descent based on a weighted sum of the three losses: %$\mathcal{L} = \lambda_s \mathcal{L}_{sup} +  \lambda_d \mathcal{L}_{dis} + \lambda_c \mathcal{L}_{con}$.
   \vspace{-0.05in}
\begin{equation}\label{eq:loss_function}
    \mathcal{L} = \lambda_s \mathcal{L}_{sup} +  \lambda_d \mathcal{L}_{dis} + \lambda_c \mathcal{L}_{con} .
       \vspace{-0.05in}
\end{equation}
After updating the student network, the dynamic teacher is updated as an exponential moving average (EMA) of the student parameters:
   $\Phi'_t = \alpha \Phi'_{t-1}  + (1-\alpha) \Phi_t$
\footnote{The subscripts of $\Phi_{B \cup N}$ and $\Phi'_{B \cup N}$ are omitted for brevity.}. $\alpha$ is a hyper-parameter to determine the amount of information taken from the student network.
At inference time, an input point cloud is directly passed to the dynamic teacher network\footnote{Both the student and dynamic teacher networks can be used for prediction during inference. We empirically found that the dynamic teacher gives better prediction results and thus use it for inference.} to predict a set of 3D bounding boxes, which are post-processed by a 3D NMS module.

\vspace{-0.1in}
\paragraph{Discussion.} Interestingly, the static teacher and the dynamic teacher are opposing each other. The conservative former is preventing the student from deviating too much from the base model, while the radical latter is pushing the student to update with new knowledge. 
Nonetheless, an equilibrium would be reached by the knowledge distilling static teacher and the consistency regularizing dynamic teacher when the co-training converges.
%On the other hand, sometimes these two teachers cooperate each other. The static teacher prevents the student from being confused by the wrongly suppressed base class objects (\textit{i.e.} background), which is accordant with the target of consistency regularization in the dynamic teacher.
%To ensure the two adversarial teachers work harmoniously together,
%avoid the model be confused by the adversarial supervision of the two teachers, 
%we adopt a ramp-up technique (\textit{c.f.} Sec.~\ref{sec:implementDetails}) to schedule the respective contributions of $\lambda_d$ and $\lambda_c$ for the losses $\mathcal{L}_{dis}$ and $\mathcal{L}_{con}$. 

\begin{table*}[t]
\centering
\scalebox{0.76}{
\begin{tabular}{|l|l |c c c | c c c | c c c|}  \hline
 &\multirow{2}{*}{\textbf{Method}} & \multicolumn{3}{c|}{$|C_{novel}|=5$}& \multicolumn{3}{c|}{$|C_{novel}|=3$}& \multicolumn{3}{c|}{$|C_{novel}|=1$}\\ \cline{3-11} 
& & \multicolumn{1}{c}{Base} & \multicolumn{1}{c}{Novel} &\multicolumn{1}{c|}{All} & \multicolumn{1}{c}{Base} & \multicolumn{1}{c}{Novel} &\multicolumn{1}{c|}{All} & \multicolumn{1}{c}{Base} & \multicolumn{1}{c}{Novel} &\multicolumn{1}{c|}{All}\\ \cline{1-10} \toprule[0.1pt]\hline
\rownumber & Base training & 57.58 & -- & -- & 53.73 & -- & -- & 55.10 & -- &  -- \\ \hline\toprule[0.1pt]\hline
\rownumber & Freeze and add  & 54.24 & 10.61  & 32.42 & 51.94 & 12.64 & 40.16 & 54.63 & 0.9  & 49.26 \\
\rownumber & Fine-tuning  & 3.48 & 54.09 & 28.79 & 4.1  & 60.17 & 20.92 & 14.86 & 1.38 & 13.51 \\
\rownumber & SDCoT w/o $\mathcal{L}_{dis}$ \& $\mathcal{L}_{con}$  & 52.17 & 50.12 & 51.14 & 38.96 & 63.68 & 46.38  & 26.83 & 24.77 & 26.63 \\
\rownumber & SDCoT w/o $\mathcal{L}_{dis}$  & 50.35 & 59.88 & 55.12 & 37.91 & 66.39 & 46.45 & 30.85  & 29.96 & 30.76 \\
\rownumber & SDCoT w/o $\mathcal{L}_{con}$ & 52.92 & 57.11 & 55.01 & 41.81 & 63.45 & 48.30 & 31.61 & 25.78 & 31.02 \\
\rownumber & \textbf{SDCoT}  & \underline{53.61} & \underline{60.80} & \underline{57.21} & \underline{44.48} & \underline{67.41} & \underline{51.36}  & \underline{36.81} & \underline{42.69} & \underline{37.40} \\ \hline\toprule[0.1pt]\hline
\rownumber & Joint training & 58.92 & 58.80 & 58.86 & 54.80 & 68.33 & 58.86 & 55.36 & 90.36 & 58.86\\ \hline\toprule[0.1pt]
\end{tabular}}
   \vspace{-0.1in}
\caption{\small{\textit{Batch incremental} 3D object detection performance (mAP@0.25) on \textbf{SUN RGB-D val} set. All the methods listed in the middle table incrementally learn on $|C_{novel}|$ novel classes. Base training is with ($10-|C_{novel}|$) base classes and joint training is with all 10 classes.}}
   \vspace{-0.1in}
 \label{tab:sunrgbd_batch}
\end{table*}

\begin{table*}[t]
\centering
\scalebox{0.76}{
\begin{tabular}{|l|l |c c c | c c c | c c c|}  \hline
&\multirow{2}{*}{\textbf{Method}} & \multicolumn{3}{c|}{$|C_{novel}|=9$}& \multicolumn{3}{c|}{$|C_{novel}|=4$}& \multicolumn{3}{c|}{$|C_{novel}|=1$}\\ \cline{3-11} 
&& \multicolumn{1}{c}{Base} & \multicolumn{1}{c}{Novel} &\multicolumn{1}{c|}{All} & \multicolumn{1}{c}{Base} & \multicolumn{1}{c}{Novel} &\multicolumn{1}{c|}{All} & \multicolumn{1}{c}{Base} & \multicolumn{1}{c}{Novel} &\multicolumn{1}{c|}{All}\\ \cline{1-10} \toprule[0.1pt]\hline
\rownumber & Base training & 60.75  & -- & -- & 53.14 & -- & -- & 56.89 & -- &  -- \\ \hline\toprule[0.1pt]\hline
\rownumber & Freeze and add  & 58.85 & 4.22  & 31.53 & 49.85 & 3.15 & 39.47 & 56.24 & 0.29 & 53.14 \\
\rownumber & Fine-tuning  & 1.91 & 52.39 & 27.15 & 1.09 & 59.44 & 14.05 & 0.25 & 12.98 & 0.96  \\
\rownumber & SDCoT w/o $\mathcal{L}_{dis}$ \& $\mathcal{L}_{con}$ & 53.09 & 46.42 & 49.76 & 48.27 & 63.87 & 51.74 & 47.91 & 27.89  & 46.80  \\
\rownumber & SDCoT w/o $\mathcal{L}_{dis}$  & 51.21 & 53.58 & 52.39 & 48.45 & 69.82 & 53.19 & 48.60 & 30.07  & 47.57 \\
\rownumber & SDCoT w/o $\mathcal{L}_{con}$ & 53.31 & 51.22 & 52.26 & 48.54  & 67.52 & 52.76 & 49.31 & 30.52 & 48.26  \\
\rownumber & \textbf{SDCoT}  & \underline{53.75} & \underline{54.91} & \underline{54.33} & \underline{49.50} & \underline{70.85} & \underline{54.25}  & \underline{52.01} & \underline{31.71} & \underline{50.89}  \\ \hline\toprule[0.1pt]\hline
\rownumber & Joint training & 58.90 & 54.13 & 56.51 & 53.16 & 68.23 & 56.51 & 57.83 & 34.16 & 56.51  \\ \hline\toprule[0.1pt]
\end{tabular}}
   \vspace{-0.1in}
\caption{\small{\textit{Batch incremental} 3D object detection performance (mAP@0.25) on \textbf{ScanNet val} set. All the methods listed in the middle table incrementally learn on $|C_{novel}|$ novel classes. Base training is with ($18-|C_{novel}|$) base classes and joint training is with all 18 classes.}}
   \vspace{-0.15in}
 \label{tab:scannet_batch}
\end{table*}

\section{Experiments}
\subsection{Datasets and Settings}
\vspace{-0.05in}
\paragraph{Datasets.} We evaluate SDCoT on the SUN RGB-D 3D object detection benchmark and ScanNet dataset. \textbf{SUN RGB-D} \cite{song2015sun} consists of 5,285 training samples and 5,050 validation samples for hundreds of object classes. To be consistent with the standard evaluation protocol in prior works (\textit{e.g.} VoteNet), we perform evaluation on the 10 most common categories. \textbf{ScanNet} \cite{dai2017scannet} consists of 1,201 training samples and 312 validation samples, where there is no amodal oriented 3D bounding boxes but point-level semantic segmentation labels. We follow VoteNet to derive the axis-aligned bounding boxes from the point-level labeling and adopt the same 18 object classes for evaluation. 
The differences between the two datasets are highlighted in the supplementary material.

%It is worth to highlight the two differences between the two datasets. First, the average number of object instances in one scene of SUN RGB-D dataset is much lower than that of ScanNet dataset (\textit{c.f.} supplementary material); since SUN RGB-D collects a scene by single-view scanning, while ScanNet reconstructs a complete scene from RGB-D video. Second, the ground-truth (GT) bounding boxes in SUN RGB-D are orientated and complete despite most objects are partially visible (see example in Figure \ref{fig:pseudo_labels} left). In contrast, the GT bounding boxes in ScanNet are axis-aligned and fit to the visible parts.

\vspace{-0.1in}
\paragraph{Setup.} To customize the datasets to the class-incremental learning setting, we take a subset of classes in alphabetical order from each dataset as $C_{base}$ and treat the remaining as $C_{novel}$, following the class splitting strategy in class-incremental image-based object detection \cite{shmelkov2017incremental}. 
$D_{base}$ is composed of training samples that contain any class of $C_{base}$ and ignores annotations for $C_{novel}$. $D_{novel}$ is constructed in a similar way. Note that $D_{base}$ and $D_{novel}$ may contain the same sample, but the annotations of this sample are different due to the change of interest on the classes. 

\vspace{-0.05in}
\paragraph{Evaluation metric.} We adopt the widely used metric in 3D point cloud object detection, \textit{i.e.} mean average precision (mAP). By default, we report mAP under 3D IoU threshold 0.25, denoted as mAP@0.25, in the following experiments.

\subsection{Implementation Details} \label{sec:implementDetails}
We set $\tau_o$ and $\tau_c$ that control the selection of pseudo labels as 0.95 and 0.9, respectively.
The weights in the loss function (\textit{i.e.} Eq. \ref{eq:loss_function}) are set as $\lambda_s$=10, $\lambda_d$=1, $\lambda_c$=10. 
We adopt a ramp-up technique \cite{tarvainen2017mean} to schedule the respective contributions of $\lambda_d$ and $\lambda_c$. Specifically, $\lambda_d$ and $\lambda_c$ ramp up from 0 to their corresponding maximum value during the first 30 epochs, using a sigmoid-shaped function $e^{-5(1-t)^2}$, where $t$ increases linearly from 0 to 1 during the ramp-up period. Following SESS, we set $\alpha$ in EMA as 0.99 during the ramp-up period and raise it to 0.999 in the following training. 
The base model $\Phi_B$ and the student network $\Phi_{B\cup N}$ are trained by an Adam optimizer. The initial learning rate for $\Phi_B$ is set to 0.001 and then decayed by 0.1 at the 80$^{th}$ and 120$^{th}$ epoch. The initial learning rate for $\Phi_{B\cup N}$ varies based on the settings of class-incremental learning.

\subsection{Baselines}\label{sec:baselines}
We design two direct and naive baselines for class-incremental 3D object detection.
The first is ``\textit{freeze and add}'': freeze the base model $\Phi_B$ that is well-trained with $D_{base}$, and then add a new classifier for $C_{novel}$ trained on $D_{novel}$ to the classifier branch of $\Phi_B$. The other is ``\textit{fine-tuning}'': fine-tune all parameters of the base model (except the old classifier) as well as a new classifier for $C_{novel}$ (randomly initialized) with $D_{novel}$. 
In addition to the two naive baselines, we also compare our SDCoT with its three variants, \textit{i.e.} without either the distillation loss ($\mathcal{L}_{dis}$) or the consistency loss ($\mathcal{L}_{con}$). Concretely, we remove the entire dynamic teacher when w/o $\mathcal{L}_{con}$ is applied; and the static teacher is just used to generate pseudo labels when w/o $\mathcal{L}_{dis}$ is applied. 
Finally, joint training that is trained on all the classes serves as the upper-bound.

\begin{table*}[t]
\centering
\scalebox{0.69}{
\begin{tabular}{|l| l|>{\centering\arraybackslash}p{1cm}>{\centering\arraybackslash}p{1cm}>{\centering\arraybackslash}p{1.3cm}>{\centering\arraybackslash}p{1cm}>{\centering\arraybackslash}p{1cm}>{\centering\arraybackslash}p{1cm}>{\centering\arraybackslash}p{1.3cm}>{\centering\arraybackslash}p{1cm}>{\centering\arraybackslash}p{1cm}>{\centering\arraybackslash}p{1cm}|c|} \hline\toprule[0.2pt]\hline
& & bathtub & bed & bookshelf & chair & desk & dresser & nightstand & sofa & table & toilet & mAP\\ \hline \toprule[0.1pt] \hline
\rownumber & B[1-5]  & 73.97 & 84.71 & 30.19 & 75.09 & 23.93 & & & & & & 57.58\\ \hline \toprule[0.1pt] \hline
\rownumber & +N[6,7,8]  & 51.57 & 84.04 & 23.83 & 62.83 & 16.94 & 26.04 & 57.34 & 59.75 & & & 47.79\\
\rownumber & +N[9,10]  & 36.59 & 79.60 & 10.35 & 60.12 & 15.16 & 12.80 & 35.15 & 56.51 & 46.95 & 88.08 & 44.13 \\ \hline \toprule[0.1pt] \hline
\rownumber & B[1-10] & 78.49 & 84.31 & 32.62 & 73.73 & 25.44 & 30.90 & 58.11 & 64.15 & 50.48 & 90.36 & 58.86 \\ \hline \toprule[0.1pt]
\end{tabular}}
    \vspace{-0.1in}
    \caption{\small{Per-class performance (AP@0.25) of SDCoT on \textbf{SUN RGB-D val} set. \textit{Setting}: sequential incremental learning of 5 novel classes. B[1-5] denotes standard training on 5 base classes. B[1-10] denotes joint training on all classes.}}
    \vspace{-0.1in}
    \label{tab:sunrgbd_sequential}
\end{table*}

\begin{table*}[t]
\centering
\scalebox{0.69}{
\begin{tabular}{|l|cccccccccccccccccc|c|} \hline\toprule[0.2pt]\hline
 & bath & bed & bkshf & cabnt & chair & cntr & curtn & desk & door & ofurn & pic & refrig & showr & sink & sofa & table & toil & wind & mAP\\ \hline \toprule[0.1pt] \hline
B[1-14]  & 75.93 & 84.17 & 47.86 & 35.73 & 87.09 & 51.50 & 44.02 & 68.67 & 45.52 & 41.47 & 6.86  & 44.08 & 60.13 & 50.97 & & & & & 53.14 \\ \hline \toprule[0.1pt] \hline
+N[15,16]  & 49.10 & 84.28 & 39.24 & 30.70 & 86.16 & 39.16 & 40.29 & 58.86 & 35.09 & 33.60 & 2.66 & 41.51 & 28.72  & 50.02 & 86.65 & 56.66 & & & 47.67 \\ 
+N[17,18]  & 39.31 & 83.22 & 37.60 & 18.62 & 82.04 & 0.39 & 30.76 & 36.78 & 21.57 & 30.48 & 0.11 & 33.38 & 27.48 & 19.70 & 84.32 & 57.18 & 95.34 & 37.73 & 40.89 \\ \hline \toprule[0.1pt]\hline
B[1-18]   & 70.85 & 85.12 & 46.70 & 37.37 & 85.79 & 54.15 & 40.83 & 66.08 & 43.17 & 41.37 & 5.84 & 50.55 & 58.62 & 57.85 & 85.22 & 55.05 & 98.50 & 34.16 & 56.51\\\hline \toprule[0.1pt]
\end{tabular}}
    \vspace{-0.1in}
    \caption{\small{Per-class performance (AP@0.25) of SDCoT on \textbf{ScanNet val} set. \textit{Setting}: sequential incremental learning of 4 novel classes. B[1-14] denotes standard training on 14 base classes. B[1-18] denotes joint training on all classes.}}
    \vspace{-0.15in}
    \label{tab:scannet_sequential}
\end{table*}

\subsection{Quantitative Results}
We evaluate the effectiveness of SDCoT in class-incremental 3D object detection task by designing two different scenarios: 1) \textit{batch incremental learning}: all the novel classes are available at once for $\Phi_{B\cup N}$ to update; and 2) \textit{sequential incremental learning}: the novel classes are split into subsets and become available sequentially. Note that the next static teacher network is updated by the current learned student network in sequential incremental learning. Furthermore, we consider different settings on the number of novel classes in batch incremental learning to eliminate the bias caused by particular classes. Specifically, we evaluate on three settings: a) $|C_{novel}| = |C_{base}|$; b) $|C_{novel}| < |C_{base}|$ and $|C_{novel}|>1$; c) $|C_{novel}|=1$. 

\paragraph{Batch incremental learning.}
Table \ref{tab:sunrgbd_batch} and \ref{tab:scannet_batch} show the comparison results of batch incremental 3D object detection performed under the three settings on SUN RGB-D and ScanNet, respectively. In each table, the upper part is a standard training on $C_{base}$, the middle part lists the results when $C_{novel}$ is incrementally added, and the bottom part is an upper-bound jointly trained on $C_{base} \cup C_{novel}$. 
As can be seen from the tables, the two naive solutions (\textit{i.e.} freeze and add, and fine-tuning) lead to extremely poor performance on either novel classes or base classes in all settings on both datasets. It is apparent that the ``freeze and add'' solution leads to sub-optimal results on $C_{novel}$, although it can largely preserves the performance on $C_{base}$. On the other hand, ``fine-tuning'' the model with new object classes leads to catastrophic forgetting of old classes. 

It is notable that incorporating pseudo labels into ground-truth labels (see 4$^{th}$ row of Table \ref{tab:sunrgbd_batch} and \ref{tab:scannet_batch}) can greatly help the incremental model preserve the knowledge from the previous classes.
Furthermore, compared to only using mixed labels, the addition of the distillation loss (see 6$^{th}$ row of Table \ref{tab:sunrgbd_batch} and \ref{tab:scannet_batch}) gains various improvements on the base classes in different settings. This shows that the distillation loss do help exploit extra knowledge from the static teacher. We also notice that the performance with $\mathcal{L}_{dis}$ surpasses that without $\mathcal{L}_{dis}$ on the novel classes in most settings. The outperformance may be due to the advantage of the distillation loss in preventing background regions from confusing the incremental model. 
When the consistency loss is added (see 5$^{th}$ row of Table \ref{tab:sunrgbd_batch} and \ref{tab:scannet_batch}), we observe consistent and significant improvements on the novel classes on all settings. The improvements show that the dynamic teacher %empowered with the perturbation scheme and consistency loss 
is very useful in learning the underlying knowledge from new data. 
Finally, despite the dataset and setting differences, our SDCoT combining the three losses (see 7$^{th}$ row of Table \ref{tab:sunrgbd_batch} and \ref{tab:scannet_batch}) achieves the best performance on both base and novel classes compared to its three variants. This clearly demonstrates the superiority of SDCoT in adapting to novel knowledge while maintaining the previous knowledge. 
It also empirically agrees with our conjecture, \textit{i.e.} the deep distillation of knowledge from the new data and base model makes the model be less susceptible to noisy and incomplete pseudo labels.

It is interesting to see that in some settings, \textit{e.g.} $|C_{novel}|=5$ on SUN RGB-D and $|C_{novel}|=9$ on ScanNet, SDCoT outperforms the upper-bound on novel classes. We attribute this outperformance to the cooperation of consistency regularization provided by the dynamic teacher and the confusion alleviation supported by the static teacher.
Another interesting finding is the large performance gap between SDCoT and the upper-bound when only the ``toilet'' class is added (\textit{i.e.} $|C_{novel}|=1$) on SUN RGB-D. This is likely due to the ``toilet'' class having very few instances (\textit{c.f.} Table~\textcolor{red}{1} in the supplementary material) in the training set, which are insufficient for the model to learn well.

\vspace{-0.1in}
\paragraph{Sequential incremental learning.}
In Table \ref{tab:sunrgbd_sequential} and \ref{tab:scannet_sequential}, we show per-class average precision (AP) of SDCoT when novel classes are added sequentially for class-incremental learning. 
% why we do not do one by one? should we explain?
We evaluate with two consecutive subsets of novel classes on SUN RGB-D and ScanNet, respectively. The incremental model adapts to the first subset of classes from the previous base model, it is subsequently treated as the base model and adapts to the second subset of classes. 
On SUN RGB-D, we achieve 44.13\% mAP on all classes (see last entry of 3$^{rd}$ row in Table \ref{tab:sunrgbd_sequential}) after adding 5 novel classes in two consecutive batches, which is lower than 57.21\% achieved by adding the 5 classes at once (see the entry at 4$^{th}$ column and 7$^{th}$ row of Table \ref{tab:sunrgbd_batch}). 
Similar pattern is found on ScanNet: the performance (\textit{i.e.} 40.89\% mAP) after sequentially adding 4 novel classes is lower than 54.25\% obtained by adding 4 classes together. 
According to the performance of each individual base class in Table \ref{tab:sunrgbd_sequential} and \ref{tab:scannet_sequential}, we find that the classes which undergoes severe performance degradation during sequential incremental learning usually have relatively poor detection ability at the beginning stage, \textit{i.e.} base training. 
Despite the performance drop of sequential incremental learning compared to batch incremental learning, it does not cause a severe catastrophic forgetting like fine-tuning.

\begin{table}[t]
\centering
\scalebox{0.8}{
\begin{tabular}{c c c | c c c}
\hline \toprule[0.4pt]
class & center & size & Base & Novel & All \\ \hline 
 \cmark & \cmark &        & 52.54 & 60.22 & 56.38  \\
 \cmark &        & \cmark &  53.57 & 60.38 & 56.98\\
 \cmark & \cmark & \cmark &  52.53 & 60.37 & 56.45 \\\hline
 \cmark &        &        & \textbf{53.61} & \textbf{60.80} & \textbf{57.21}   \\
\hline \toprule[0.6pt]
\end{tabular}}
\vspace{-0.1in}
\caption{\small{Effects of different distillation targets under the setting of $|C_{novel}|=5$ on SUN RGB-D dataset.}}
\label{tab:distill_target}
\vspace{-0.2in}
\end{table}

\subsection{Design Choices of Distillation Loss} \label{sec:design_choice}
\begin{figure}[t]
\centering
\includegraphics[scale=0.6]{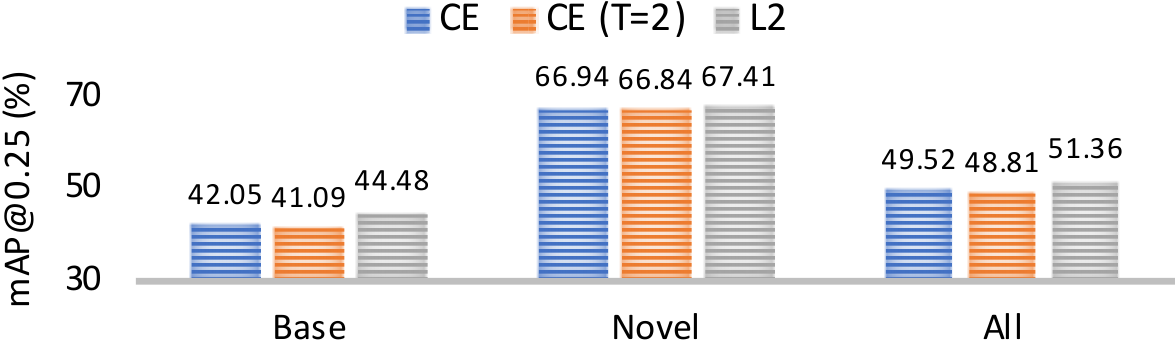}
   \vspace{-0.1in}
   \caption{\small{Effects of different distillation loss functions under the setting of $|C_{novel}|=3$ on SUN RGB-D dataset.}}
\label{fig:how_to_distill}
\vspace{-0.1in}
\end{figure}

We investigate the effects of various designs of the distillation loss. More specifically, we study different distillation targets (\textit{i.e.} classification logit, bounding box regression values including \textit{center} and \textit{size}) and alternative loss functions (\textit{i.e.} cross-entropy and knowledge distillation losses).  
\vspace{-0.05in}
\paragraph{What to distill?} Table \ref{tab:distill_target} summarizes the effects of using different distilled targets when computing the final distillation loss. Note that we compute the mean square error between the corresponding outputs from $\Phi_B$ and $\Phi_{B \cup N}$ for size- and center-aware distillation losses, in addition to our original class-aware distillation loss. As can be seen from the table, the size- and center-aware distillation are unable to extract more useful information from the previous knowledge. In fact, they slightly harm the performance on the base classes in the given setting. Consequently, we only distill knowledge from the classification logits.
\vspace{-0.1in}
\paragraph{How to distill?} To evaluate the effects of different loss functions, we replace the L2 norm loss in Eq.~\ref{eq:distillation_loss} with cross-entropy loss and knowledge distillation loss \cite{hinton2015distilling} that is an cross-entropy loss with temperature, respectively. Figure \ref{fig:how_to_distill} shows that the L2 norm loss is a better choice for class-incremental 3D object detection.

\begin{figure}[t]
\centering
\includegraphics[scale=0.38]{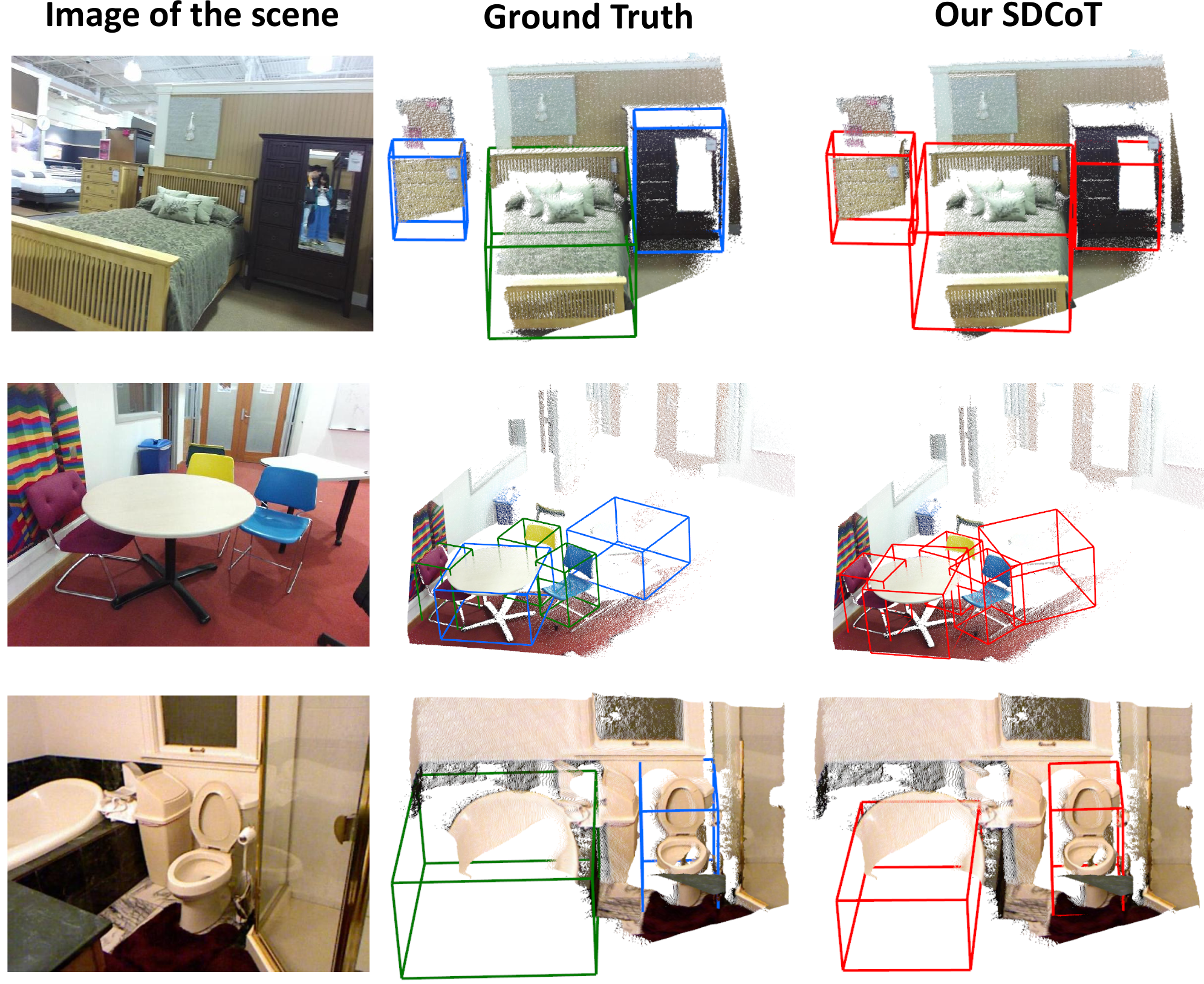}
   \vspace{-0.1in}
   \caption{\small{Qualitative results on \textbf{SUN RGB-D val} set. \textcolor{green}{Green} and \textcolor{blue}{Blue} represent GT annotations w.r.t $C_{base}$ and $C_{novel}$, respectively.}}
   \vspace{-0.1in}
\label{fig:qualitative_sunrgbd}
\end{figure}

\begin{figure}[t]
\centering
\includegraphics[scale=0.32]{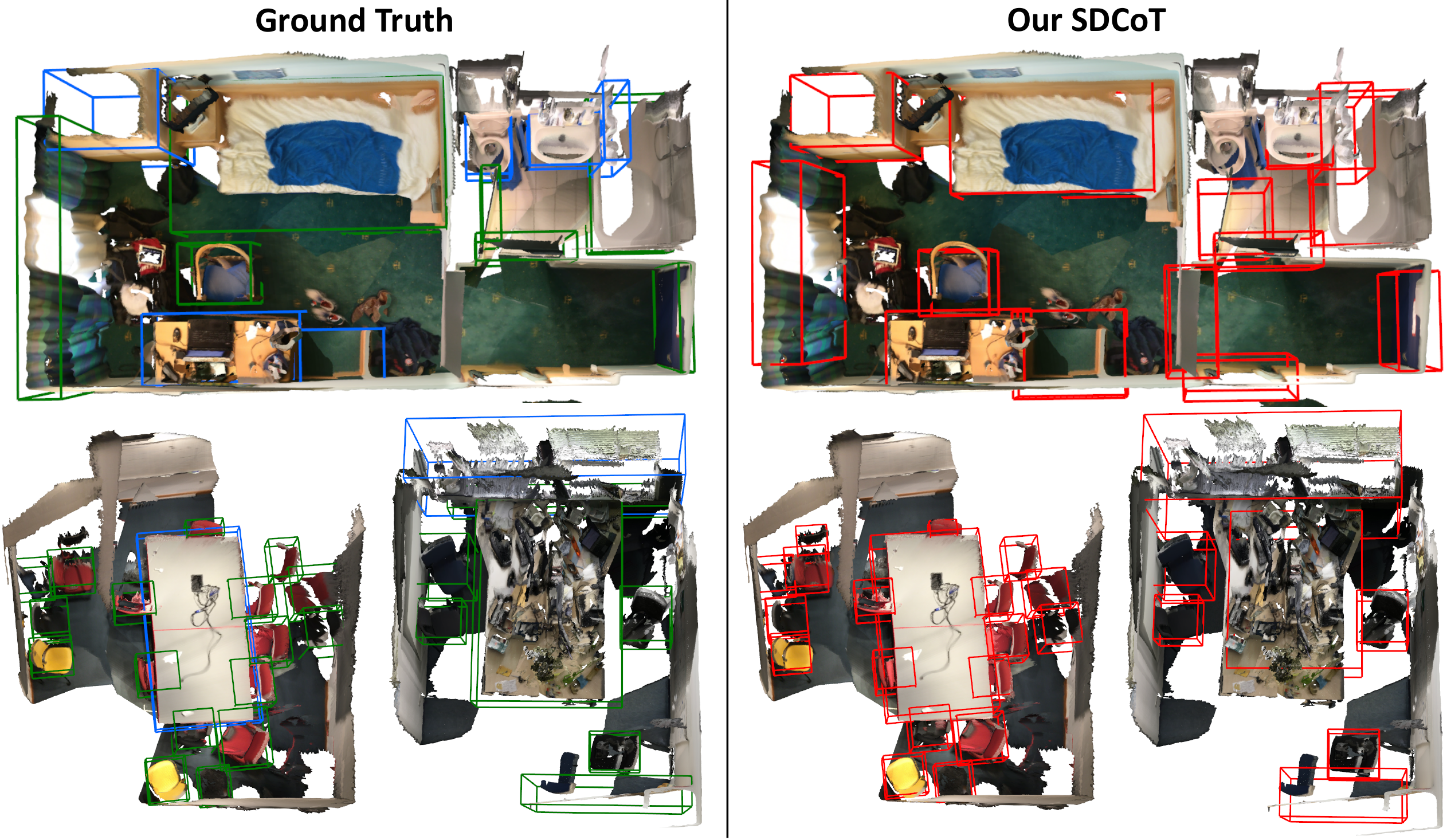}
   \vspace{-0.1in}
   \caption{\small{Qualitative results on \textbf{ScanNet val} set. 
   %\textcolor{green}{Green} and \textcolor{blue}{Blue} bboxes are GT annotations w.r.t $C_{base}$ and $C_{novel}$, respectively.
   }}
\label{fig:qualitative_scannet}
\vspace{-0.1in}
\end{figure}

\subsection{Qualitative Results}
Figure \ref{fig:qualitative_sunrgbd} and \ref{fig:qualitative_scannet} show the qualitative results of our SDCoT on SUN RGB-D and ScanNet, respectively. 
Despite the very challenging (\textit{e.g.} partially visible objects and cluttered scenes) and diverse (\textit{e.g.} bedroom, bathroom, and conference room) scenes, our SDCoT is able to nicely detect the novel classes as well as greatly retain the detection capacity on the base classes in all these scenes. In addition, we provide some failure examples in the supplementary material.

\begin{figure}[t]
\centering
\includegraphics[scale=0.45]{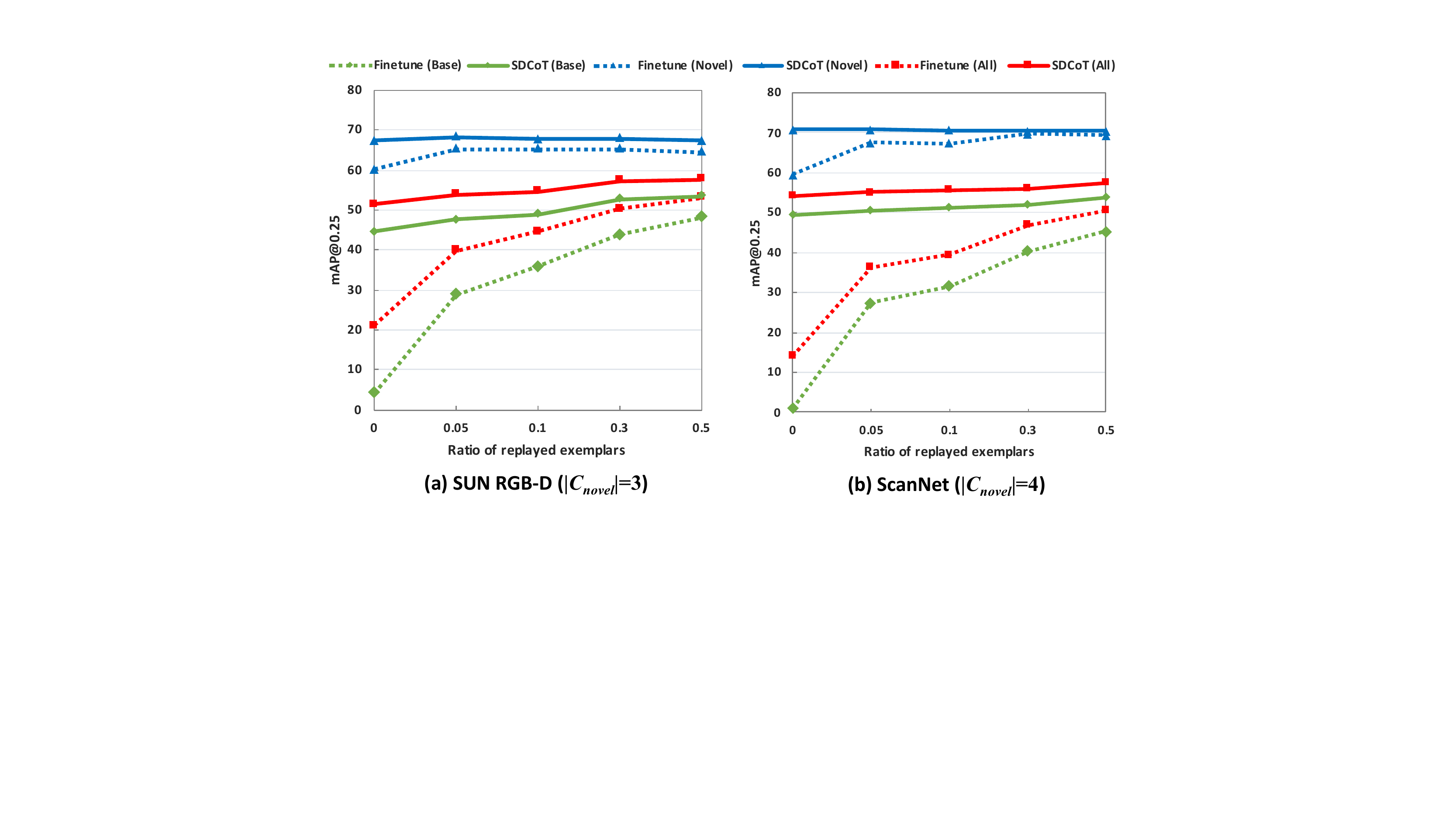}
   \vspace{-0.1in}
   \caption{\small{Comparison with fine-tuning baseline on SUN RGB-D and ScanNet val sets with varying ratios of old data. \textit{Setting}: batch incremental 3D object detection with $|C_{novel}|$ novel classes.}}
   	\vspace{-0.1in}
\label{fig:with_replay}
\end{figure}

\subsection{Compatibility with Replayed Exemplars}
In the class-incremental learning of image classification task, it is common to store a small set of samples from old data (\textit{i.e.} exemplars) to prevent catastrophic forgetting. However, the amount of its contribution in the class-incremental 3D object detection task is unclear.
We adopt the simplest but effective strategy, \textit{i.e.} random sampling \cite{chaudhry2018riemannian}, to select exemplars from old data\footnote{We ensure that all base classes are present in the exemplars, or otherwise we re-sample until the condition is met.}. Interestingly, our SDCoT can easily incorporate these exemplars into the ``mixed labels" as labeled instances without any change to the framework. To demonstrate the effects of different number of replayed exemplars in class-incremental 3D object detection, we sample different ratios of old data and compare the results with the baseline method (\textit{i.e.} fine-tuning) on the two datasets, as shown in Figure \ref{fig:with_replay}. As can be seen, when more replayed exemplars are added, fine-tuning baseline achieves significant improvements on base classes while our SDCoT only gets very slight improvements. This indicates that our method is capable of persevering old knowledge, which makes it less sensitive to the addition of replayed exemplars. 
Furthermore, it can be seen that our SDCoT consistently outperforms fine-tuning over all percentages of replaying (\textit{c.f.} the supplementary material for the numerical comparisons). 

\section{Conclusion}
This paper studies the new and practical problem of class-incremental 3D object detection. To this end, we proposed SDCoT: an effective static-dynamic co-teaching method to incrementally detect novel classes without revisiting any previous training sample.
Our SDCoT greatly addresses the catastrophic forgetting issue and further helps the model adapt to the novel classes. We demonstrated the effectiveness SDCoT over a variety of class-incremental 3D object detection scenarios on SUN RGB-D and ScanNet datasets. 
We hope that our study serves as a motivation for future works on this practical problem.

\section{Acknowledgement}
This research is supported in part by the National Research Foundation, Singapore under its AI Singapore Program (AISG Award No: AISG2-RP-2020-016) and the Tier 2 grant MOET2EP20120-0011 from the Singapore Ministry of Education.

\appendix
\section{Supplementary Material}
In this appendix, we provide some observations on benchmark datasets, more details on our backbone such as architecture and loss function, per-class evaluation under batch incremental learning, some failure cases, and numerical comparison results with replayed exemplars.

\subsection{Observations on Benchmark Datasets} 
It is worth to highlight the two differences between the two datasets. First, the average number of object instances in one scene of SUN RGB-D dataset is much lower than that of ScanNet dataset, since SUN RGB-D collects a scene by single-view scanning, while ScanNet reconstructs a complete scene from RGB-D video. This difference can be observed in our qualitative examples (\textit{c.f.} Figure \textcolor{red}{6} and \textcolor{red}{7} in the main paper), where the scenes in ScanNet are more cluttered. Second, the ground-truth (GT) bounding boxes in SUN RGB-D are orientated and complete despite most objects are partially visible (see example in Figure \textcolor{red}{3} left in the main paper). In contrast, the GT bounding boxes in ScanNet are axis-aligned and fit to the visible parts.

Table \ref{tbl:sunrgbd-statistics} and \ref{tbl:scannet-statistics} list the class names in alphabetical order and the statistics of per-class instance number as well as scan number (\textit{i.e.} the number of scans containing the corresponding class) on training set of SUN RGB-D and ScanNet, respectively. As can be see from the two tables, the two datasets are both highly unbalanced across classes. This unbalanced data can cause the insufficient training problem when the added novel class has very few samples, \textit{e.g.} the addition of `toilet' class with only 174 training samples does not perform well under batch incremental 3D object detection setting (see the results under $|C_{novel}|=1$ setting in Table \textcolor{red}{1} of the main paper).
%Furthermore, we can observe that the number of instances per scan in ScanNet is larger than that in SUN RGB-D. 

\begin{table*}[t]
	\centering
	%\begin{center}
	\scalebox{0.85}{
		\begin{tabular}{l|c c c c c c c c c c c c c}
			\hline\toprule[0.2pt]
			\textbf{class name}& \textbf{bathtub} & \textbf{bed} &\textbf{bookshelf} & \textbf{chair}& \textbf{desk}& \textbf{dresser}& \textbf{night stand}& \textbf{sofa}& \textbf{table}& \textbf{toilet}  \\\hline
			\textbf{\# scans} & 73 & 640 & 166 & 2,499 & 682 & 155 & 254 & 515 & 1,772 & 174 \\\hline
			\textbf{\# instances}  & 73 & 795 & 212 & 9,499 & 1,028 & 193 & 302 & 724 & 2,671 & 177 \\\hline\toprule[0.2pt]
		\end{tabular}
	}
	\vspace{-0.1in}
	\caption{Statistics of per-class instance number and scan number on \textbf{SUN RGB-D train} set.}
	\label{tbl:sunrgbd-statistics}
\end{table*}

\begin{table*}[t]
	\centering
	%\begin{center}
	\scalebox{0.78}{
		\begin{tabular}{p{2cm} | p{1.8cm} p{1.2cm} p{2cm} p{1.4cm} p{1.3cm} p{1.4cm} p{1.4cm} p{1.2cm} p{1.1cm} }
			\hline\toprule[0.2pt]
			\textbf{class name} & \textbf{bathtub} & \textbf{bed} &\textbf{bookshelf} & \textbf{cabinet}& \textbf{chair}& \textbf{counter}& \textbf{curtain}& \textbf{desk}& \textbf{door}   \\\hline
			\textbf{\# scans}  & 113 & 245 & 172 & 564 & 798 & 179 & 193 & 327 & 874  \\\hline
			\textbf{\# instances}  & 113 & 307 & 300 & 1,427 & 4,357 & 216 & 292 & 551 & 2,026  \\\hline\toprule[0.2pt]
	\end{tabular}}
	\scalebox{0.78}{
		\begin{tabular}{p{2cm}| p{2.3cm} p{1.2cm} p{1.8cm} p{2.5cm} p{0.9cm} p{0.9cm} p{1cm} p{1cm} p{1.2cm}}
			\hline\toprule[0.2pt]
			\textbf{class name}  &\textbf{other furniture} & \textbf{picture} & \textbf{refrigerator} & \textbf{shower curtain} & \textbf{sink} & \textbf{sofa} & \textbf{table} & \textbf{toilet} & \textbf{window}  \\\hline
			\textbf{\# scans} & 890 & 321 & 177 & 106 & 322 & 279 & 623 & 189 & 609  \\\hline
			\textbf{\# instances} & 1,985 & 661 & 186 & 116 & 390 & 406 & 1,271 & 201 & 928 \\\hline\toprule[0.2pt]
	\end{tabular}}
\vspace{-0.1in}
	\caption{Statistics of per-class instance number and scan number on \textbf{ScanNet train} set.}
	\label{tbl:scannet-statistics}
\end{table*}

\subsection{More Details on Our Backbone}\label{sec: arch_details}
\paragraph{Architecture Details.} Our backbone, \textit{i.e.} modified VoteNet, is comprised of four modules (as shown in Figure \textcolor{red}{2} of the main paper): 1) feature learning backbone that sub-samples $M$ points (\textit{i.e.} \textit{seeds}) from $N$ input points with enriched features via PointNet++ \cite{qi2017pointnet++}; 2) voting module that generates expected centers of objects (\textit{i.e.} \textit{votes}) from seeds via several MLP layers; 3) proposal generator that samples $K$ votes and groups a set of neighboring votes for each sampled votes to form vote clusters, which are subsequently passed to a light PointNet network to generate $K$ object proposals with aggregated features (\textit{i.e.} proposal features); 4) predicting module including a regressor and a classifier, which takes proposal features as input and yields prediction scores of box parameters (\textit{i.e.} 2 objectness scores, 3 center offsets, 3 size offsets, and 2$NH$ heading scores) and box category ($NC$ category scores), respectively. 
In SUN RGB-D, $NH=12$; in ScanNet, $NH=1$. $NC$ is dynamically updated according to the setting of class-incremental learning, \textit{i.e.} how many novel classes to be added. 

The architectures of the feature learning backbone and the voting modules are the same as the original VoteNet \cite{qi2019deep}. The proposal generator module is a set abstraction (SA) layer followed by two MLP layers with output sizes of 128, 128, which is same as the proposal module in VoteNet but without the last MLP layer. The regressor in predicting module is a Linear layer with output size of (2+3+3+2$NH$), while the classifier is a Linear layer without a bias. 

\paragraph{Supervised Loss Function.} Similar to VoteNet, the supervised loss $\mathcal{L}_{sup}$ is a weighted sum of a voting loss, an objectiveness loss, a 3D bounding box estimation loss and a semantic classification loss:
\begin{equation}
	\mathcal{L}_{sup} = \mathcal{L}_{vote-reg} + \lambda_1 \mathcal{L}_{obj-cls} + \lambda_2 \mathcal{L}_{box} + \lambda_3 \mathcal{L}_{sem-cls}.
\end{equation}
We set $\lambda_1=0.5$, $\lambda_2=1$, and $\lambda_3=0.2$. The vote regression loss $\mathcal{L}_{vote-reg}$, objectness loss $\mathcal{L}_{obj-cls}$, and semantic classification loss $\mathcal{L}_{sem-cls}$ are the same as defined in VoteNet. Following VoteNet, the box loss $\mathcal{L}_{box}$ is computed as:
\begin{equation}
	\begin{split}
		\mathcal{L}_{box} = & \mathcal{L}_{center-reg} + 0.1 \mathcal{L}_{angle-cls} + \mathcal{L}_{angle-reg} ~~~~\\
		& + \mathcal{L}_{size-reg},
	\end{split}
\end{equation}
with the exception that the size classification loss is removed, as we replace the class-aware size prediction with class-agnostic one. We refer the readers to VoteNet for more details on the computation of each loss term.

\subsection{Per-class Evaluation}\label{sec:per_class_evaluation}
In this section, we report per-class performance (Average Precision with 0.25 box IoU) comparison results of batch incremental 3D object detection performed under the three settings on SUN RGB-D and ScanNet, respectively.

\paragraph{SUN RGB-D.} Table \ref{tab:sunrgbd_batch_5novel}, \ref{tab:sunrgbd_batch_3novel} and \ref{tab:sunrgbd_batch_1novel} list the per-class comparison results of batch incremental 3D object detection when $|C_{novel}|=5$, $|C_{novel}|=3$ and $|C_{novel}|=1$, respectively.  
\paragraph{ScanNet.} Table \ref{tab:scannet_batch_9novel}, \ref{tab:scannet_batch_4novel} and \ref{tab:scannet_batch_1novel} list the per-class comparison results of batch incremental 3D object detection when $|C_{novel}|=9$, $|C_{novel}|=4$ and $|C_{novel}|=1$, respectively.

From the per-class performance under different batch incremental settings on both datasets, we can see that our SDCoT is able to achieve optimal results on novel classes compared to the ``Freeze and add'' baseline, and largely prevent the model from catastrophic forgetting of base classes compared to the ``Fine-tuning'' baseline. 
Furthermore, compared to its three variants, our SDCoT achieves the best performance in most classes and show comparable performance in the remaining classes. 

\begin{table*}[t]
	%\begin{center}
	\centering
	\scalebox{0.76}{
		\begin{tabular}{|l| l|>{\centering\arraybackslash}p{1cm}>{\centering\arraybackslash}p{1cm}>{\centering\arraybackslash}p{1.3cm}>{\centering\arraybackslash}p{1cm}>{\centering\arraybackslash}p{1cm}>{\centering\arraybackslash}p{1cm}>{\centering\arraybackslash}p{1.3cm}>{\centering\arraybackslash}p{1cm}>{\centering\arraybackslash}p{1cm}>{\centering\arraybackslash}p{1cm}|c|} \hline
			& Method & bathtub & bed & bookshelf & chair & desk & dresser & nightstand & sofa & table & toilet & mAP\\ \hline \toprule[0.1pt] \hline
			\rownumber & Base training  & 73.97 & 84.71 & 30.19 & 75.09 & 23.93 & & & & & & 57.58\\ \hline \toprule[0.1pt] \hline
			\rownumber & Freeze and add & 72.86  &74.34  & 30.32 & 70.29  & 23.38  & 0.56 & 0.09  & 10.25  & 12.04 & 30.11 & 32.42   \\\hline
			\rownumber & Fine-tuning &  0.02 & 2.64 & 0.45 & 1.95  & 12.36  & 21.37  & 58.16  & 57.16  & 47.84 & 85.95 & 28.79  \\\hline\toprule[0.1pt] \hline
			\rownumber & SDCoT w/o $\mathcal{L}_{dis}$ \& $\mathcal{L}_{con}$ &  68.63 & \textbf{83.30} & 23.56 & 68.63  & 16.72  & 20.43  & 40.25  & 57.17  & 47.33 & 85.43 & 51.14  \\\hline
			\rownumber & SDCoT w/o $\mathcal{L}_{dis}$ & 69.35  & 80.07 & 16.66 & 69.51  & 16.23  & 34.28  & 61.42  & 62.32  & 50.94 & 90.45 & 55.12  \\\hline
			\rownumber & SDCoT w/o $\mathcal{L}_{con}$ & 73.09  & 81.29 & \textbf{23.78} & 68.87  & 17.54  & 33.79  &  53.80 & 61.06  & 48.12 & 88.78 & 55.01  \\\hline
			\rownumber & \textbf{SDCoT}  & \textbf{75.41}  & 82.15 & 22.13 & \textbf{70.58}  & \textbf{17.81}  & \textbf{35.56}  & \textbf{61.80}  &  \textbf{62.98} & \textbf{52.99} & \textbf{90.67} & \textbf{57.21}  \\\hline \toprule[0.1pt] \hline
			\rownumber & Joint training & 78.49 & 84.31 & 32.62 & 73.73 & 25.44 & 30.90 & 58.11 & 64.15 & 50.48 & 90.36 & 58.86 \\ \hline \toprule[0.1pt]
	\end{tabular}}
	%\end{center}
	\vspace{-0.1in}
	\caption{\small{Per-class performance (AP@0.25) comparison on \textbf{SUN RGB-D val} set. \textit{Setting}: batch incremental learning of \textbf{5 novel classes}.}}
	\label{tab:sunrgbd_batch_5novel}
\end{table*}

\begin{table*}[h]
	%\begin{center}
	\centering
	\scalebox{0.76}{
		\begin{tabular}{|l| l|>{\centering\arraybackslash}p{1cm}>{\centering\arraybackslash}p{1cm}>{\centering\arraybackslash}p{1.3cm}>{\centering\arraybackslash}p{1cm}>{\centering\arraybackslash}p{1cm}>{\centering\arraybackslash}p{1cm}>{\centering\arraybackslash}p{1.3cm}>{\centering\arraybackslash}p{1cm}>{\centering\arraybackslash}p{1cm}>{\centering\arraybackslash}p{1cm}|c|} \hline
			& Method & bathtub & bed & bookshelf & chair & desk & dresser & nightstand & sofa & table & toilet & mAP\\ \hline \toprule[0.1pt] \hline
			\rownumber & Base training  & 72.81  & 83.55  & 30.83 & 75.14  & 26.56  & 27.87 & 59.36 & & & & 53.73 \\ \hline \toprule[0.1pt] \hline
			\rownumber & Freeze and add &  74.85 & 76.28 & 31.02 & 72.60 & 24.04 & 27.93  & 56.89  & 11.95  & 13.43  & 12.56  & 40.16  \\\hline
			\rownumber & Fine-tuning & 0.52  & 16.47 & 0.34 & 2.34  & 7.99  & 0.40  & 0.61  & 56.16  & 48.76 & 75.61 & 20.92  \\\hline\toprule[0.1pt] \hline
			\rownumber & SDCoT w/o $\mathcal{L}_{dis}$ \& $\mathcal{L}_{con}$ & 71.46  & 77.28 & 16.21 & 67.32  & 11.59  & 0.24  & 28.64  & 61.18  & 46.46 & 83.40 & 46.38  \\\hline
			\rownumber & SDCoT w/o $\mathcal{L}_{dis}$ & \textbf{74.70}  & 73.28 & 12.11 & 65.97  & 12.67  & 0.31  & 26.30  & 60.73  & 49.39 & \textbf{89.05} & 46.45  \\\hline
			\rownumber & SDCoT w/o $\mathcal{L}_{con}$ &  71.28 & 77.36 & \textbf{20.17} & 67.60  & 13.87  &  7.32 & 35.08  & 61.65  & 47.28 & 81.43 & 48.30  \\\hline
			\rownumber & \textbf{SDCoT}  & 70.61  &  \textbf{78.10} & 18.82 & \textbf{68.09}  & \textbf{18.28}  & \textbf{13.43}  & \textbf{44.00}  & \textbf{62.99}  & \textbf{50.29} & 88.94 & \textbf{51.36}  \\\hline \toprule[0.1pt] \hline
			\rownumber & Joint training & 78.49 & 84.31 & 32.62 & 73.73 & 25.44 & 30.90 & 58.11 & 64.15 & 50.48 & 90.36 & 58.86 \\ \hline \toprule[0.1pt]
	\end{tabular}}
	%\end{center}
	\vspace{-0.1in}
	\caption{\small{Per-class performance (AP@0.25) comparison on \textbf{SUN RGB-D val} set. \textit{Setting}: batch incremental learning of \textbf{3 novel classes}.}}
	\label{tab:sunrgbd_batch_3novel}
\end{table*}

\begin{table*}[h]
	%\begin{center}
	\centering
	\scalebox{0.76}{
		\begin{tabular}{|l| l|>{\centering\arraybackslash}p{1cm}>{\centering\arraybackslash}p{1cm}>{\centering\arraybackslash}p{1.3cm}>{\centering\arraybackslash}p{1cm}>{\centering\arraybackslash}p{1cm}>{\centering\arraybackslash}p{1cm}>{\centering\arraybackslash}p{1.3cm}>{\centering\arraybackslash}p{1cm}>{\centering\arraybackslash}p{1cm}>{\centering\arraybackslash}p{1cm}|c|} \hline
			& Method & bathtub & bed & bookshelf & chair & desk & dresser & nightstand & sofa & table & toilet & mAP\\ \hline \toprule[0.1pt] \hline
			\rownumber & Base training  & 71.77  & 85.01  & 33.19 & 73.76  & 25.73  & 29.41 & 61.31 & 64.79 & 50.91 & & 55.10  \\ \hline \toprule[0.1pt] \hline
			\rownumber & Freeze and add & 67.33  & 84.69  & 32.83 & 73.56  & 25.11  & 29.59  & 63.06  & 65.08  & 50.43 & 0.90 & 49.26  \\\hline
			\rownumber & Fine-tuning & 2.54  & 17.72 & 0.82 & 53.81  & 5.99  & 1.46  & 11.21  & 17.86  & 22.28 & 1.38 & 13.51  \\\hline \toprule[0.1pt] \hline
			\rownumber & SDCoT w/o $\mathcal{L}_{dis}$ \& $\mathcal{L}_{con}$ & 70.38  & 28.06 & 5.60 & 58.11  & 10.22  & 2.42  & 3.72  & 30.25  & 32.72 & 24.77 & 26.63  \\\hline
			\rownumber & SDCoT w/o $\mathcal{L}_{dis}$ &  67.32 & \textbf{54.57} & 7.28 & 59.28  & 7.31  & 1.73  & 7.05  & \textbf{47.09}  & 26.03 & 29.96 & 30.76  \\\hline
			\rownumber & SDCoT w/o $\mathcal{L}_{con}$ & 71.82  & 26.03 & 9.63 & 60.64  & 11.93  & \textbf{8.58}  & \textbf{28.95}  & 30.39  & 36.47 & 25.78 & 31.02  \\\hline
			\rownumber & \textbf{SDCoT}  & \textbf{72.63}  & 52.04  & \textbf{10.92} & \textbf{64.55}  & \textbf{14.99}  &  4.65 & 28.26  & 45.58  & \textbf{37.71} & \textbf{42.69} & \textbf{37.40}  \\\hline \toprule[0.1pt] \hline
			\rownumber & Joint training & 78.49 & 84.31 & 32.62 & 73.73 & 25.44 & 30.90 & 58.11 & 64.15 & 50.48 & 90.36 & 58.86 \\ \hline \toprule[0.1pt]
	\end{tabular}}
	%\end{center}
	\vspace{-0.1in}
	\caption{\small{Per-class performance (AP@0.25) comparison on \textbf{SUN RGB-D val} set. \textit{Setting}: batch incremental learning of \textbf{1 novel classes}.}}
	\label{tab:sunrgbd_batch_1novel}
\end{table*}

\begin{table*}[t]
	%\begin{center}
	\centering
	\scalebox{0.64}{
		\begin{tabular}{|l|l|cccccccccccccccccc|c|} \hline\toprule[0.2pt]\hline
			& Method & bath & bed & bkshf & cabnt & chair & cntr & curtn & desk & door & ofurn & pic & refrig & showr & sink & sofa & table & toil & wind & mAP\\ \hline \toprule[0.1pt] \hline
			\rownumber & Base training  & 77.91 & 85.15 & 47.95 & 40.36 & 88.17 & 57.80 & 35.69 & 69.38 & 44.34 &  &  &  &  &  &  &  &  &  & 60.75  \\ \hline 
			\rownumber & Freeze and add  & 74.76 & 83.83 & 43.95 & 42.93 & 87.04 & 49.58 & 36.27 & 66.62 & 44.66 & 1.03 & 0 & 10.30 & 4.62 & 1.21 & 9.28 & 7.45 & 1.75 & 2.31 &  31.53   \\ \hline 
			\rownumber & Fine-tuning  & 0.01 & 2.66 & 0.94 & 2.45 & 1.26 & 0.13 & 1.17 & 8.05 & 0.51 & 35.22 & 6.02 & 36.06 & 58.27 & 65.51 & 86.64 & 58.36 & 95.01 & 30.44 & 27.15  \\ \hline \toprule[0.1pt] \hline
			\rownumber & SDCoT w/o $\mathcal{L}_{dis}$ \& $\mathcal{L}_{con}$  & 53.36 & 83.87 & 37.45 & \textbf{32.52} & 85.13 & \textbf{49.43} & 37.91 & 63.19 & 34.97 & 29.18 & 4.17 & 44.98 & 45.58 & 34.22 & 86.03 & 57.46 & 93.14 & 23.08 & 49.76  \\ \hline
			\rownumber & SDCoT w/o $\mathcal{L}_{dis}$  & 54.15 & \textbf{85.50} & 39.95 & 27.29 & \textbf{86.22} & 37.53 & 36.03 & 64.91 & 29.33 & 40.69 & \textbf{9.41} & 45.59 & \textbf{61.22} & 48.48 & 85.65 & \textbf{64.02} & 96.66 & 30.45 & 52.39  \\ \hline 
			\rownumber & SDCoT w/o $\mathcal{L}_{con}$  & 55.40 & 82.77 & \textbf{42.79} & 29.90 & 84.67 & 45.50 & \textbf{38.02} & 62.43 & \textbf{38.28} & 39.93 & 3.51 & 39.26 & 58.34 & 47.92 & 87.13 & 57.66 & 95.57 & \textbf{31.68} & 52.26  \\ \hline 
			\rownumber & \textbf{SDCoT}  & \textbf{57.10} & 82.61 & 42.59 & 30.44 & 86.16 & 45.77 & 37.16 & \textbf{65.26} & 36.67 & \textbf{42.14} & 6.12 & \textbf{51.48} & 61.06 & \textbf{53.11} & \textbf{89.07} & 63.05 & \textbf{98.30} & 29.87 & \textbf{54.33}  \\ \hline \toprule[0.1pt] \hline
			\rownumber & Joint training  &70.85 & 85.12 & 46.70 & 37.37 & 85.79 & 54.15 & 40.83 & 66.08 & 43.17 & 41.37 & 5.84 & 50.55 & 58.62 & 57.85 & 85.22 & 55.05 & 98.50 & 34.16 & 56.51  \\\hline \toprule[0.1pt]
	\end{tabular}}
	%\end{center}
	\vspace{-0.1in}
	\caption{\small{Per-class performance (AP@0.25) comparison on \textbf{ScanNet val} set. \textit{Setting}: batch incremental learning of \textbf{9 novel classes}.}}
	\label{tab:scannet_batch_9novel}
\end{table*}

\begin{table*}[t]
	%\begin{center}
	\centering
	\scalebox{0.64}{
		\begin{tabular}{|l|l|cccccccccccccccccc|c|} \hline\toprule[0.2pt]\hline
			& Method & bath & bed & bkshf & cabnt & chair & cntr & curtn & desk & door & ofurn & pic & refrig & showr & sink & sofa & table & toil & wind & mAP\\ \hline 
			\rownumber & Base training  & 75.93 & 84.17 & 47.86 & 35.73 & 87.09 & 51.50 & 44.02 & 68.67 & 45.52 & 41.47 & 6.86 & 44.08 & 60.13 & 50.97 &  &  &  &  & 53.14 \\ \hline \toprule[0.1pt] \hline
			\rownumber & Freeze and add  & 76.98 & 86.03 & 40.37 & 32.28 & 86.35 & 60.24 & 41.91 & 53.50 & 25.38 & 38.32 & 3.41 & 44.94 & 49.12 & 59.02 & 1.37 & 5.52 & 4.99 & 0.70 & 39.47  \\ \hline 
			\rownumber & Fine-tuning  & 0.02 & 6.67 & 0.21 & 0.52 & 1.05 & 0.06 & 0.55 & 4.44 & 0.89 & 0.56 & 0.0 & 0.17 & 0.03 & 0.0 & 65.08 & 54.94 & 85.42 & 32.32 & 14.05  \\ \hline \toprule[0.1pt] \hline
			\rownumber & SDCoT w/o $\mathcal{L}_{dis}$ \& $\mathcal{L}_{con}$  & 67.10 & \textbf{85.16} & 37.88 & \textbf{29.72} & 84.58 & 49.63 & 35.22 & \textbf{64.72} & \textbf{40.36} & 35.54 & 4.64 & 38.91 & 52.76 & 49.56 & 83.91 & 53.03 & 92.94 & 25.62 & 51.74  \\ \hline
			\rownumber & SDCoT w/o $\mathcal{L}_{dis}$  & \textbf{77.04} & 84.62 & 40.32 & 26.56 & 83.77 & \textbf{49.71} & \textbf{41.35} & 58.50 & 36.61 & 36.34 & 1.34 & 40.49 & 61.66 & 39.92 & 85.26 & 60.13 & \textbf{97.68} & 36.22 & 53.19  \\ \hline 
			\rownumber & SDCoT w/o $\mathcal{L}_{con}$  & 72.58 & 83.04 & 37.60 & 27.16 & \textbf{84.76} & 42.94 & 35.26 & 60.95 & 38.00 & 38.32 & 4.89 & \textbf{43.24} & 59.39 & \textbf{51.48} & 85.24 & 59.47 & 93.53 & 31.84 & 52.76  \\ \hline 
			\rownumber & \textbf{SDCoT}  & 75.76 & 84.16 & \textbf{43.08} & 28.20 & 84.34 & 43.05 & 38.07 & 60.79 & 39.55 & \textbf{39.91} & \textbf{4.93} & 40.33 & \textbf{63.79} & 47.12 & \textbf{87.35} & \textbf{62.68} & 96.82 & \textbf{36.57} & \textbf{54.25}  \\ \hline \toprule[0.1pt] \hline
			\rownumber & Joint training  & 70.85 & 85.12 & 46.70 & 37.37 & 85.79 & 54.15 & 40.83 & 66.08 & 43.17 & 41.37 & 5.84 & 50.55 & 58.62 & 57.85 & 85.22 & 55.05 & 98.50 & 34.16 & 56.51  \\\hline \toprule[0.1pt]
	\end{tabular}}
	%\end{center}
	\vspace{-0.1in}
	\caption{\small{Per-class performance (AP@0.25) comparison on \textbf{ScanNet val} set. \textit{Setting}: batch incremental learning of \textbf{4 novel classes}.}}
	\label{tab:scannet_batch_4novel}
\end{table*}

\begin{table*}[t]
	%\begin{center}
	\centering
	\scalebox{0.64}{
		\begin{tabular}{|l|l|cccccccccccccccccc|c|} \hline\toprule[0.2pt]\hline
			& Method & bath & bed & bkshf & cabnt & chair & cntr & curtn & desk & door & ofurn & pic & refrig & showr & sink & sofa & table & toil & wind & mAP\\ \hline 
			\rownumber & Base training  & 72.07 & 84.19 & 46.61 & 38.40 & 86.41 & 51.63 & 37.82 & 71.05 & 41.36 & 37.61 & 4.76 & 50.79 & 47.53 & 58.50 & 87.01 & 58.19 & 93.23 &  & 56.89  \\ \hline \toprule[0.1pt] \hline
			\rownumber & Freeze and add  & 73.12 & 85.56 & 44.65 & 37.88 & 86.43 & 48.04 & 36.05 & 69.25 & 40.83 & 38.22 & 4.77 & 47.79 & 48.07 & 55.99 & 87.72 & 57.05 & 94.74 & 0.29 & 53.14  \\ \hline 
			\rownumber & Fine-tuning  & 0.0 & 0.0 & 0.18 & 0.08 & 0.0 & 0.0 & 2.26 & 0.0 & 1.60 & 0.0 & 0.0 & 0.0 & 0.11 & 0.0 & 0.0 & 0.03 & 0.0 & 12.98 &  0.96 \\ \hline \toprule[0.1pt] \hline
			\rownumber & SDCoT w/o $\mathcal{L}_{dis}$ \& $\mathcal{L}_{con}$  & \textbf{74.42} & 79.72 & 31.75 & 24.89 & 81.88 & 33.60 & 32.51 & 63.42 & 28.75 & 26.91 & 0.0 & 34.28 & 45.37 & 31.85 & 79.81 & 51.95 & \textbf{93.33} & 27.89 & 46.80  \\ \hline
			\rownumber & SDCoT w/o $\mathcal{L}_{dis}$  & 52.84 & 82.49 & 41.32 & 28.79 & 83.81 & \textbf{39.34} & \textbf{33.84} & 64.01 & 35.98 & 26.55 & 0.06 & 40.07 & 31.11 & 42.26 & 83.71 & 52.80 & 87.19 & 30.07 & 47.57  \\ \hline 
			\rownumber & SDCoT w/o $\mathcal{L}_{con}$  & 68.51 & 80.69 & 36.15 & 28.83 & 84.16 & 25.37 & 13.32 & 66.29 & 35.14 & 27.94 & \textbf{0.84} & \textbf{47.06} & \textbf{47.57} & 43.41 & \textbf{86.10} & 55.49 & 91.44 & 30.52 & 48.26  \\ \hline 
			\rownumber & \textbf{SDCoT}  & 69.18 & \textbf{83.63} & \textbf{44.29} & \textbf{30.64} & \textbf{85.36} & 36.51 & 31.03 & \textbf{66.99} & \textbf{36.31} & \textbf{30.88} & 0.13 & 46.02 & 44.27 & \textbf{46.55} & 84.08 & \textbf{56.36} & 92.00 & \textbf{31.71} & \textbf{50.89}  \\ \hline \toprule[0.1pt] \hline
			\rownumber & Joint training  &  70.85 & 85.12 & 46.70 & 37.37 & 85.79 & 54.15 & 40.83 & 66.08 & 43.17 & 41.37 & 5.84 & 50.55 & 58.62 & 57.85 & 85.22 & 55.05 & 98.50 & 34.16 & 56.51  \\\hline \toprule[0.1pt]
	\end{tabular}}
	%\end{center}
	\vspace{-0.1in}
	\caption{\small{Per-class performance (AP@0.25) comparison on \textbf{ScanNet val} set. \textit{Setting}: batch incremental learning of \textbf{1 novel classes}.}}
	\label{tab:scannet_batch_1novel}
\end{table*}

\subsection{Failure Cases} \label{sec:failure_cases}
\begin{figure}[t]
	\centering
	\includegraphics[scale=0.385]{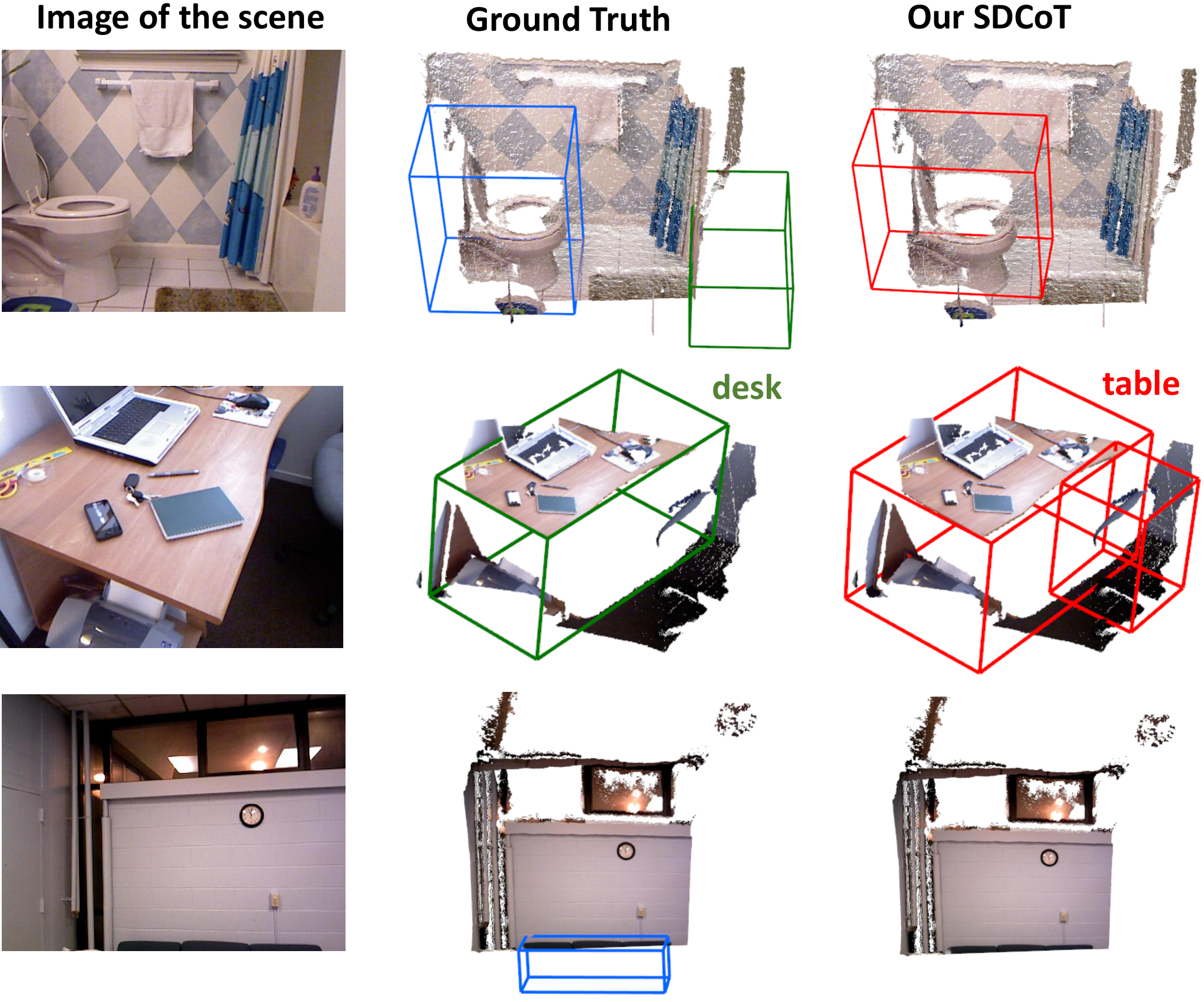}
	\caption{\small{Failure cases from \textbf{SUN RGB-D val} set. \textcolor{green}{Green} and \textcolor{blue}{Blue} bboxes are GT annotations w.r.t $C_{base}$ and $C_{novel}$, respectively. We show three examples from top to bottom.
			%Our SDCoT fails to detect `bathtub' in the first scene and mistakenly detects `desk' as `table' in the second scene.
	}}
	\label{fig:failurecases_sunrgbd}
\end{figure}

In Figure \ref{fig:failurecases_sunrgbd}, we show three failure examples on SUN RGB-D dataset. As can be seen from the first example (1$^{st}$ row in Figure \ref{fig:failurecases_sunrgbd}), our SDCoT fails to detect the bathtub (\textit{i.e.} base class) that is extremely truncated. Nonetheless, it is interesting to see that our SDCoT is able to detect the chair (\textit{i.e.} base class) in the second example (2$^{nd}$ row in Figure \ref{fig:failurecases_sunrgbd}) despite only a small portion of it is visible and ground-truth annotation is not available. We suspect the difference in the detection capacity between the two classes (\textit{i.e.} `bathtub' and `chair') is caused by the 
imbalanced number of training samples. The `chair' class has much more training examples than the `bathtub' class (compare 4$^{th}$ column with 1$^{st}$ column in Table \ref{tbl:sunrgbd-statistics}). In the second example, our SDCoT mistakenly detects the desk (\textit{i.e.} base class) as table (\textit{i.e.} novel class). This is likely due to the very similar geometric structures of these two classes; and the learned model is prone to recognize such geometric structure as the class that is recently seen and with more training samples (\textit{i.e.} `table'). In the last example, we show an extremely challenging scenario, where our SDCoT fails to detect the sofa that is totally invisible.

\begin{figure*}[t]
	\centering
	\includegraphics[scale=0.4]{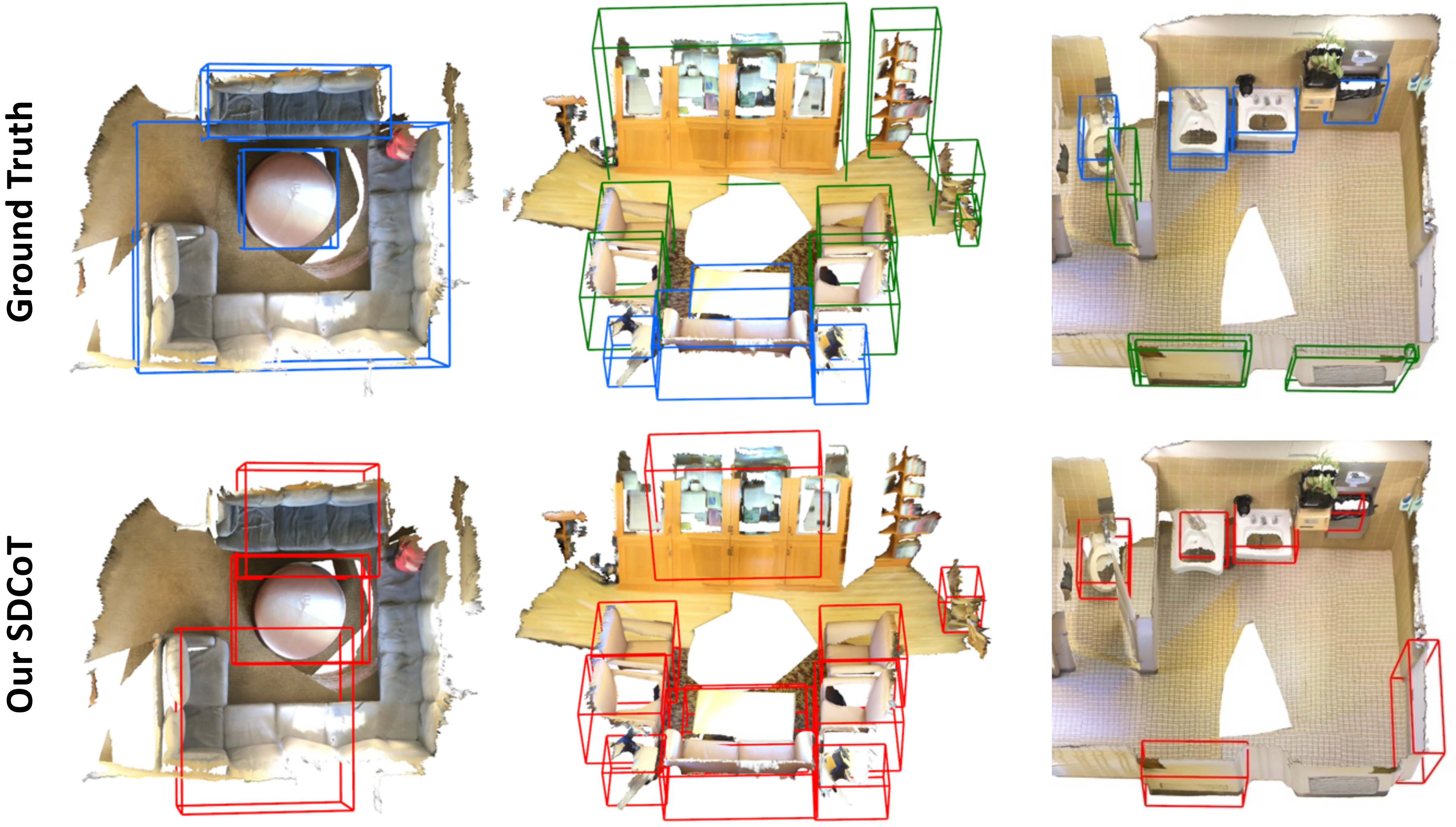}
	\caption{Failure cases from \textbf{ScanNet val} set. \textcolor{green}{Green} and \textcolor{blue}{Blue} bboxes are GT annotations w.r.t $C_{base}$ and $C_{novel}$, respectively. We show three examples from left to right.}
	\label{fig:failurecases_scannet}
\end{figure*}

We also show three failure examples on ScanNet dataset in Figure \ref{fig:failurecases_scannet}. We use these examples to illustrate several common reasons of failure in our SDCoT: 1) fail to detect thin objects such as door and picture (see the missed doors in the bottom right of the third example); 2) fail to detect partly visible objects (see the missed bookshelf in the upper right and missed chair in the middle right of the second example); 
%3) failure to detect objects rarely seen during training (see the missed `otherfurniture' in the middle right area of the second example); 
3) hard to detect the entire objects with larger shapes (see the sofa in the lower part of the first example and the bookshelf in the upper part of the second example); and 4) wrongly detect background areas as objects (see the red bounding box on the right wall in the third example, where our SDCoT mistakenly detects the wall part as a door).

\subsection{Numerical Results with Replayed Exemplars} \label{sec:replay_exemplars}
In Table \ref{tab:replay_sunrgbd} and \ref{tab:replay_scannet}, we list the numerical results of comparisons between fine-tuning baseline and our SDCoT on the two datasets with varying ratios of replayed exemplars, for a clear illustration. We also list the exact number of the selected replayed exemplars under different ratios on both datasets.
%Each sub-sampled point cloud (with 50k points) of one scene in ScanNet is 1.2MB, while the corresponding point cloud in SUN RGB-D is around 800KB. 

\begin{table*}[t]
	\centering
	\scalebox{0.9}{
		\begin{tabular}{|l |c c c | c c c | c c c| c c c|}  \hline
			\multirow{2}{*}{\textbf{Method}} & \multicolumn{3}{c|}{5\% (167)}& \multicolumn{3}{c|}{10\% (334)}& \multicolumn{3}{c|}{30\% (1,002)} & \multicolumn{3}{c|}{50\% (1,671)}\\ \cline{2-13} 
			&\multicolumn{1}{c}{Base} & \multicolumn{1}{c}{Novel} &\multicolumn{1}{c|}{All} & \multicolumn{1}{c}{Base} & \multicolumn{1}{c}{Novel} &\multicolumn{1}{c|}{All} & \multicolumn{1}{c}{Base} & \multicolumn{1}{c}{Novel} &\multicolumn{1}{c|}{All} & \multicolumn{1}{c}{Base} & \multicolumn{1}{c}{Novel} &\multicolumn{1}{c|}{All}\\ \cline{1-13} \toprule[0.1pt]\hline
			Fine-tuning  & 28.83 & 65.27 & 39.76  & 35.78 & 65.19 & 44.60 & 43.84 & 65.22 & 50.25  & 48.19 & 64.49 & 53.08\\
			\textbf{SDCoT}  & \textbf{47.62} & \textbf{68.29} & \textbf{53.82} & \textbf{48.96} & \textbf{67.67} & \textbf{54.57}  & \textbf{52.65} & \textbf{67.84} &  \textbf{57.21} & \textbf{53.53} & \textbf{67.24} & \textbf{57.65} \\ \hline\toprule[0.1pt]
	\end{tabular}}
\vspace{-0.1in}
	\caption{\small{Comparison with fine-tuning baseline on \textbf{SUN RGB-D val} dataset with varying ratios of old data. \textit{Setting}: batch incremental 3D object detection of 3 novel classes (\textit{i.e.} $|C_{novel}|=3$). Note that the number associated with the percentage indicates the number of replayed exemplars.}}
	\label{tab:replay_sunrgbd}
\end{table*}

\begin{table*}[t]
	\centering
	\scalebox{0.9}{
		\begin{tabular}{|l |c c c | c c c | c c c| c c c|}  \hline
			\multirow{2}{*}{\textbf{Method}} & \multicolumn{3}{c|}{5\% (59)}& \multicolumn{3}{c|}{10\% (118)}& \multicolumn{3}{c|}{30\% (356)} & \multicolumn{3}{c|}{50\% (594)}\\ \cline{2-13} 
			&\multicolumn{1}{c}{Base} & \multicolumn{1}{c}{Novel} &\multicolumn{1}{c|}{All} & \multicolumn{1}{c}{Base} & \multicolumn{1}{c}{Novel} &\multicolumn{1}{c|}{All} & \multicolumn{1}{c}{Base} & \multicolumn{1}{c}{Novel} &\multicolumn{1}{c|}{All} & \multicolumn{1}{c}{Base} & \multicolumn{1}{c}{Novel} &\multicolumn{1}{c|}{All}\\ \cline{1-13} \toprule[0.1pt]\hline
			Fine-tuning  & 27.41 & 67.42 & 36.30  & 31.53 & 67.35 & 39.49 & 40.37 & 69.59 & 46.86  & 45.26 & 69.39 & 50.62\\
			\textbf{SDCoT}  & \textbf{50.54} & \textbf{70.81} & \textbf{55.04} & \textbf{51.29} & \textbf{70.60} & \textbf{55.58}  & \textbf{51.92} & \textbf{70.36} &  \textbf{56.02} & \textbf{53.74} & \textbf{70.39} & \textbf{57.44} \\ \hline\toprule[0.1pt]
	\end{tabular}}
\vspace{-0.1in}
	\caption{\small{Comparison with fine-tuning baseline on \textbf{ScanNet val} dataset with varying ratios of old data. \textit{Setting}: batch incremental 3D object detection of 4 novel classes (\textit{i.e.} $|C_{novel}|=4$). Note that the number associated with the percentage indicates the number of replayed exemplars.}}
	\label{tab:replay_scannet}
\end{table*}

% Use \bibliography{yourbibfile} instead or the References section will not appear in your paper
%\clearpage
\bibliography{egbib}
% \nobibliography{aaai22}

\end{document}